\documentclass{article}

\usepackage[preprint, nonatbib]{neurips_2024}

\usepackage[utf8]{inputenc} % allow utf-8 input
\usepackage[T1]{fontenc}    % use 8-bit T1 fonts
\usepackage{hyperref}       % hyperlinks
\usepackage{url}            % simple URL typesetting
\usepackage{booktabs}       % professional-quality tables
\usepackage{amsfonts}       % blackboard math symbols
\usepackage{nicefrac}       % compact symbols for 1/2, etc.
\usepackage{microtype}      % microtypography
\usepackage{xcolor}         % colors
\usepackage{amsmath, amssymb}
\usepackage{amsthm}
\usepackage{tikz}
\usepackage{subfigure}
\usepackage{multirow}
\usepackage{wrapfig}
\usepackage{authblk}
\usepackage{tabularx}

\usepackage[ruled]{algorithm2e}
\usepackage{algpseudocode}

\usepackage[titletoc, title]{appendix}
\usepackage{titletoc}

\hypersetup{
    colorlinks=true,  % 设置为true使链接颜色生效，而不是用颜色方框
    linkcolor=blue,   % 内部链接颜色（例如，目录到章节的链接）
    filecolor=magenta, % 文件链接的颜色
    urlcolor=cyan,    % 外部URL的颜色
    citecolor=green   % 引用链接的颜色
}

\usetikzlibrary{arrows.meta,decorations.pathreplacing, patterns, intersections, calc}

\newtheorem{proposition}{Proposition}
\newtheorem{theorem}{Theorem}
\newtheorem{corollary}{Corollary}
\newtheorem{lemma}{Lemma}
\newtheorem{definition}{Definition}

\title{Accelerating Non-Maximum Suppression:\par A Graph Theory Perspective}

\author[1*]{\textbf{King-Siong Si}}
\author[1*]{\textbf{Lu Sun}}
\author[1$\dag$]{\textbf{Weizhan Zhang}}
\author[2]{\textbf{Tieliang Gong}}
\author[2]{\authorcr\textbf{Jiahao Wang}}

\author[3]{\textbf{Jiang Liu}}
\author[3]{\textbf{Hao Sun}}

\affil[1]{School of Computer Science and Technology, MOEKLINNS Lab, Xi'an Jiaotong University}
\affil[2]{School of Computer Science and Technology, BDKE Lab, Xi'an Jiaotong University}
\affil[3]{Institute of Artificial Intelligence (TeleAI), China Telecom}

\affil[ ]{\texttt{\{sjsinx, sunlu.cs\}@stu.xjtu.edu.cn, \{zhangwzh, gongtl\}@xjtu.edu.cn}}
\affil[ ]{\texttt{uguisu@stu.xjtu.edu.cn, \{black\_liu\_1025, sun.010\}@163.com}}

\begin{document}

\maketitle

\def\thefootnote{*}\footnotetext{Equal contribution}
\def\thefootnote{$\dag$}\footnotetext{Corresponding author}

\begin{abstract}
Non-maximum suppression (NMS) is an indispensable post-processing step in object detection. With the continuous optimization of network models, NMS has become the ``last mile'' to enhance the efficiency of object detection. This paper systematically analyzes NMS from a graph theory perspective for the first time, revealing its intrinsic structure. Consequently, we propose two optimization methods, namely QSI-NMS and BOE-NMS. The former is a fast recursive divide-and-conquer algorithm with negligible mAP loss, and its extended version (eQSI-NMS) achieves optimal complexity of $\mathcal{O}(n\log n)$. The latter, concentrating on the locality of NMS, achieves an optimization at a constant level without an mAP loss penalty. Moreover, to facilitate rapid evaluation of NMS methods for researchers, we introduce NMS-Bench, the first benchmark designed to comprehensively assess various NMS methods. Taking the YOLOv8-N model on MS COCO 2017 as the benchmark setup, our method QSI-NMS provides $6.2\times$ speed of original NMS on the benchmark, with a $0.1\%$ decrease in mAP. The optimal eQSI-NMS, with only a $0.3\%$ mAP decrease, achieves $10.7\times$ speed. Meanwhile, BOE-NMS exhibits $5.1\times$ speed with no compromise in mAP.
\end{abstract}

% Finally, we introduced NMS-Bench, the first project enabling researchers to easily and quickly validate NMS methods and compare them with benchmark NMS methods.

\section{Introduction}\label{intro}

Object detection is a highly significant and popular topic in computer vision, widely applied in various domains, e.g., multiple object tracking \cite{mot, xu2019deep, 1421760}, medical imaging analysis \cite{mia, shen2017deep}, multimodal object detection \cite{Cao_2023_CVPR, Zhang_2024_CVPR}, and autonomous driving \cite{autodriving, dollar2009pedestrian, geiger2013vision}. In recent years, there has been significant attention on the real-time performance of object detection, with notable successes achieved in several research endeavors \cite{FasterRCNN, YOLO, ssd}. Non-maximum suppression (NMS) \cite{greedynms_human_dectection} is a post-processing technique used to eliminate duplicate detection boxes and obtain final detections. Some research on NMS has indeed enhanced the mean average precision (mAP) of object detection, but they have also introduced additional computational overhead. 

Currently, most CNN-based object detection models (such as the R-CNN family \cite{rcnn, fastrcnn, FasterRCNN} and the YOLO series \cite{YOLO, yolov2, yolov3,yolov4}) consist of two parts in the testing phase: model inference and post-processing. In recent years, with the continuous emergence of model lightweighting techniques \cite{hinton2015distilling, wang2020cspnet, ding2021repvgg}, the time cost of model inference has been significantly reduced. As a result, NMS gradually becomes a bottleneck in the pipeline of object detection \cite{detrbeatsyolo}. To address this, some studies \cite{yoloact, ClusterNMS, wang2020solov2} have proposed parallelization methods to enhance NMS efficiency. However, these methods do not reduce computational overhead; they rely heavily on efficient parallel computing to reduce overall time costs. The degree of parallelism depends on the hardware environment (such as processor type, quantity, and number of cores) and architecture, leading to significant variations in efficiency when models are deployed across different platforms. Additionally, NMS research lacks a unified evaluation framework for two main reasons. First, existing NMS methods require a complete model inference for each test, consuming a significant amount of unnecessary computational resources. Second, different NMS methods are tested on different platforms using various models, making comparisons between different NMS algorithms challenging.

To reduce the computational overhead of NMS, we first map the set of bounding boxes obtained from model inference to a graph $\mathcal{G}$. We then conduct a comprehensive and systematic analysis of the intrinsic structure of NMS from a graph theory perspective. Each box is considered a node in the graph, and the suppression relationships are represented as arcs. We discovered that this forms a directed acyclic graph (DAG), allowing us to solve NMS using dynamic programming. This indicates that as long as the graph $\mathcal{G}$ can be quickly constructed, NMS can be efficiently performed. Through the analysis of $\mathcal{G}$, we find that it contains many weakly connected components (WCCs), and most of them are small. Based on these two characteristics, we propose two optimization strategies. First, due to the nature of dynamic programming, different WCCs are independent. We can use a divide-and-conquer algorithm to break down the problem into smaller subproblems corresponding to these WCCs and solve them recursively. Inspired by quicksort, we propose quicksort induced NMS (QSI-NMS), which provides $6.18\times$ speed with a negligible $0.1\%$ decrease in mAP compared to original NMS in YOLOv8-N \cite{yolov8} on MS COCO 2017 \cite{ms-coco}. Furthermore, by analyzing the structure of QSI-NMS, we propose extended QSI-NMS (eQSI-NMS) with a complexity of $\mathcal{O}(n \log n)$, achieving state-of-the-art performance. Second, leveraging the locality suppression characteristic of NMS, where most weakly connected components are small, we exclude boxes that cannot have suppression relationships through geometric analysis. This led to the development of boxes outside excluded NMS (BOE-NMS), which provides $5.12\times$ speed with no compromise in mAP compared to original NMS in YOLOv8-N on MS COCO 2017.

To facilitate the evaluation and comparison of NMS algorithms, we introduce NMS-Bench, the first end-to-end benchmark for rapid NMS validation. By decoupling model inference and post-processing, we save substantial computational resources, enabling NMS validation to be completed within minutes. Moreover, by implementing NMS algorithms fairly within this framework, different NMS algorithms can be compared on an equal footing. Thus, we integrate data, benchmarking methods, and evaluation metrics into a single framework, enabling end-to-end rapid validation and simplifying NMS research for researchers.

In summary, our contributions are as follows:
\begin{itemize}
    \item We present the first comprehensive analysis of the NMS algorithm from a graph theory perspective, uncovering the intrinsic structure of NMS;
    \item We propose two efficient NMS algorithms based on the properties of the NMS-induced graph;
    \item We introduce NMS-Bench, the first end-to-end benchmark for rapid NMS validation.
\end{itemize}

\section{Problem Definition}\label{sec:pro-def}

Original NMS, employs the intersection over union (IOU) between bounding boxes as the criterion for mutual suppression. Specifically, Given a set of candidate bounding boxes $\mathcal{B}$, original NMS selects the box $b^*$ with the highest confidence score from $\mathcal{B}$, removes it from $\mathcal{B}$, and adds it to the final output set $\mathcal{D}$. Then, it computes IOUs between $b^*$ and all other boxes in $\mathcal{B}$. If the IOU with a certain box $b$ is greater than a given threshold $N_t$, then $b$ is removed from $\mathcal{B}$. This process is repeated until $\mathcal{B}$ is empty.

In general, NMS during post-processing is an algorithm, which takes a list of detection bounding boxes $\mathcal{B}$ with corresponding confidence scores $\mathcal{S}$ as input, and outputs a subset $\mathcal{D}$ of $\mathcal{B}$. And for convenience, we denote the cardinality of $\mathcal{B}$ by $n$, i.e., $n=\vert\mathcal{B}\vert$. Formally, the NMS algorithm takes $(\mathcal{B},\mathcal{S})$ as input, and outputs a sequence $K=(k_1, k_2,\ldots, k_n)$, where
\[
    \begin{cases}
        k_i=1 &\text{if }b_i\in\mathcal{D};\\
        k_i=0 &\text{otherwise}.
    \end{cases}
\]  

% Formally, the NMS algorithm implements a mapping: $(\mathcal{B},\mathcal{S})\mapsto\mathcal{K}$, $\mathcal{K}=\{k_i\vert i=1,2,\ldots,n\}$, where
% \[
%     \begin{cases}
%         k_i=1 &\text{if }b_i\in\mathcal{D};\\
%         k_i=0 &\text{otherwise}.
%     \end{cases}
% \]  
And an evaluation function is a mapping $\mathit{e}: \{0,1\}^{n} \mapsto \mathbb{R}$, where a larger value of $\mathit{e}$ indicates a better NMS. The goal of our research is to enhance algorithm efficiency under the condition that 
\[
\mathit{e}(K_{origin})-\mathit{e}(K)<\varepsilon, 
\]
where $\varepsilon>0$ represents the tolerance factor and $K_{origin}$ is the output of original NMS algorithm. In the object detection tasks of this paper, we use mAP as the evaluation function $e$, and we set $\varepsilon$ to $1\%$.

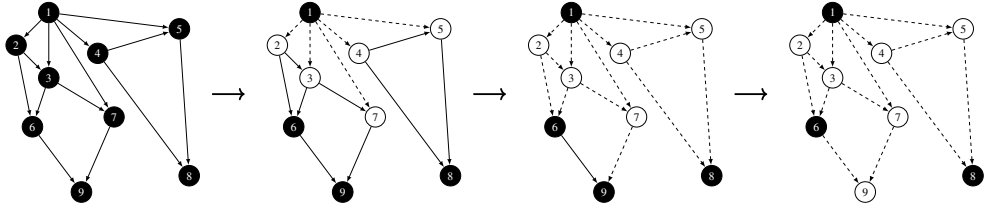
\begin{figure}[ht]
    \centering
    % \documentclass[tikz, convert, convert={outext=.pdf,
% 		command=\unexpanded{}}]{standalone}
% \usetikzlibrary{arrows.meta,decorations.pathreplacing, patterns, intersections, calc}
% \usepackage{fontspec}
% \setmainfont{Times New Roman}

% \begin{document}

\resizebox{\linewidth}{!}{
\begin{tikzpicture}
    \definecolor{node1_c}{HTML}{000000}
    \definecolor{node0_c}{HTML}{FFFFFF}

    \tikzset{
        point/.style = {
            draw, minimum size=.5cm, circle, thick
        },
        arrow/.style = {
            thick, -latex, 
        },
        dash-arrow/.style = {
            thick, -latex, dashed
        }
    }

    \draw [draw=none] (0, 0) grid (32, 8);

    \node (a0) [point, fill=node1_c, text=white] at (2, 6.5) {1};

    \node (b0) [point, fill=node1_c, text=white] at (1, 5.5) {2};
    \node (b1) [point, fill=node1_c, text=white] at (2, 4.5) {3};
    \node (b2) [point, fill=node1_c, text=white] at (3.5, 5.25) {4};
    \node (b3) [point, fill=node1_c, text=white] at (6, 6) {5};

    \node (c0) [point, fill=node1_c, text=white] at (1.5, 3) {6};
    \node (c1) [point, fill=node1_c, text=white] at (4, 3.3) {7};
    \node (c2) [point, fill=node1_c, text=white] at (6.3, 1.5) {8};

    \node (d0) [point, fill=node1_c, text=white] at (3, 1) {9};

    \draw [arrow] (a0) -- (b0);
    \draw [arrow] (a0) -- (b1);
    \draw [arrow] (a0) -- (b2);
    \draw [arrow] (a0) -- (b3);
    \draw [arrow] (a0) -- (c1);

    \draw [arrow] (b0) -- (b1);
    % \draw [arrow] (b2) -- (c1);
    \draw [arrow] (b2) -- (b3);

    \draw [arrow] (b0) -- (c0);
    \draw [arrow] (b1) -- (c0);
    \draw [arrow] (b1) -- (c1);
    \draw [arrow] (b2) -- (c2);
    \draw [arrow] (b3) -- (c2);

    \draw [arrow] (c0) -- (d0);
    \draw [arrow] (c1) -- (d0);

    \node (a01) [point, fill=node1_c, text=white] at (10, 6.5) {1};

    \node (b01) [point, fill=node0_c, text=black] at (9, 5.5) {2};
    \node (b11) [point, fill=node0_c, text=black] at (10, 4.5) {3};
    \node (b21) [point, fill=node0_c, text=black] at (11.5, 5.25) {4};
    \node (b31) [point, fill=node0_c, text=black] at (14, 6) {5};

    \node (c01) [point, fill=node1_c, text=white] at (9.5, 3) {6};
    \node (c11) [point, fill=node0_c, text=black] at (12, 3.3) {7};
    \node (c21) [point, fill=node1_c, text=white] at (14.3, 1.5) {8};

    \node (d01) [point, fill=node1_c, text=white] at (11, 1) {9};

    \draw [dash-arrow] (a01) -- (b01);
    \draw [dash-arrow] (a01) -- (b11);
    \draw [dash-arrow] (a01) -- (b21);
    \draw [dash-arrow] (a01) -- (b31);
    \draw [dash-arrow] (a01) -- (c11);

    \draw [arrow] (b01) -- (b11);
    % \draw [arrow] (b21) -- (c11);
    \draw [arrow] (b21) -- (b31);

    \draw [arrow] (b01) -- (c01);
    \draw [arrow] (b11) -- (c01);
    \draw [arrow] (b11) -- (c11);
    \draw [arrow] (b21) -- (c21);
    \draw [arrow] (b31) -- (c21);

    \draw [arrow] (c01) -- (d01);
    \draw [arrow] (c11) -- (d01);

    \node (a02) [point, fill=node1_c, text=white] at ($(a01) + (8, 0)$) {1};

    \node (b02) [point, fill=node0_c, text=black] at ($(b01) + (8, 0)$) {2};
    \node (b12) [point, fill=node0_c, text=black] at ($(b11) + (8, 0)$) {3};
    \node (b22) [point, fill=node0_c, text=black] at ($(b21) + (8, 0)$) {4};
    \node (b32) [point, fill=node0_c, text=black] at ($(b31) + (8, 0)$) {5};

    \node (c02) [point, fill=node1_c, text=white] at ($(c01) + (8, 0)$) {6};
    \node (c12) [point, fill=node0_c, text=black] at ($(c11) + (8, 0)$) {7};
    \node (c22) [point, fill=node1_c, text=white] at ($(c21) + (8, 0)$) {8};

    \node (d02) [point, fill=node1_c, text=white] at ($(d01) + (8, 0)$) {9};

    \draw [dash-arrow] (a02) -- (b02);
    \draw [dash-arrow] (a02) -- (b12);
    \draw [dash-arrow] (a02) -- (b22);
    \draw [dash-arrow] (a02) -- (b32);
    \draw [dash-arrow] (a02) -- (c12);

    \draw [dash-arrow] (b02) -- (b12);
    % \draw [dash-arrow] (b22) -- (c12);
    \draw [dash-arrow] (b22) -- (b32);

    \draw [dash-arrow] (b02) -- (c02);
    \draw [dash-arrow] (b12) -- (c02);
    \draw [dash-arrow] (b12) -- (c12);
    \draw [dash-arrow] (b22) -- (c22);
    \draw [dash-arrow] (b32) -- (c22);

    \draw [arrow] (c02) -- (d02);
    \draw [dash-arrow] (c12) -- (d02);

    \node (a03) [point, fill=node1_c, text=white] at ($(a02) + (8, 0)$) {1};

    \node (b03) [point, fill=node0_c, text=black] at ($(b02) + (8, 0)$) {2};
    \node (b13) [point, fill=node0_c, text=black] at ($(b12) + (8, 0)$) {3};
    \node (b23) [point, fill=node0_c, text=black] at ($(b22) + (8, 0)$) {4};
    \node (b33) [point, fill=node0_c, text=black] at ($(b32) + (8, 0)$) {5};

    \node (c03) [point, fill=node1_c, text=white] at ($(c02) + (8, 0)$) {6};
    \node (c13) [point, fill=node0_c, text=black] at ($(c12) + (8, 0)$) {7};
    \node (c23) [point, fill=node1_c, text=white] at ($(c22) + (8, 0)$) {8};

    \node (d03) [point, fill=node0_c, text=black] at ($(d02) + (8, 0)$) {9};

    \draw [dash-arrow] (a03) -- (b03);
    \draw [dash-arrow] (a03) -- (b13);
    \draw [dash-arrow] (a03) -- (b23);
    \draw [dash-arrow] (a03) -- (b33);
    \draw [dash-arrow] (a03) -- (c13);

    \draw [dash-arrow] (b03) -- (b13);
    % \draw [dash-arrow] (b23) -- (c13);
    \draw [dash-arrow] (b23) -- (b33);

    \draw [dash-arrow] (b03) -- (c03);
    \draw [dash-arrow] (b13) -- (c03);
    \draw [dash-arrow] (b13) -- (c13);
    \draw [dash-arrow] (b23) -- (c23);
    \draw [dash-arrow] (b33) -- (c23);

    \draw [dash-arrow] (c03) -- (d03);
    \draw [dash-arrow] (c13) -- (d03);

    \draw [ultra thick, ->] (7, 4) -- (8, 4);
    \draw [ultra thick, ->] (15, 4) -- (16, 4);
    \draw [ultra thick, ->] (23, 4) -- (24, 4);

\end{tikzpicture}
}
% \end{document}
    \caption{Dynamic programming in topological sorting. The color of the node represents the $\mathit{\delta}$ value, i.e., black represents $1$, and white represents $0$. Before suppression, each node is black. In topological sorting, traversed arcs are represented by dashed lines, showing they have been removed from the graph. After the topological sorting is completed, we can find that nodes $1$, $6$, and $8$ are all black, that is, the last boxes retained are $b_1$, $b_6$, and $b_8$.}
    \label{fig:methodology:1}
\end{figure}

% Therefore, there are two main research directions for NMS,
% \begin{itemize}
%     \item to improve $\textit{e}(\mathcal{K})$, e.g., improving average precision (AP) in object detection tasks; 
%     \item to improve algorithm efficiency under the condition that $\mathit{e}(\mathcal{K}_{origin})-\mathit{e}(\mathcal{K})<\varepsilon$, where $\varepsilon>0$ represents the tolerance factor and $\mathcal{K}_{origin}$ is the output of the original NMS algorithm, i.e., Greedy NMS.
% \end{itemize}
% Our research mainly focuses on the second one, to improve efficiency of NMS.

%%% add more details in the start of methodology
% In this section, we will analyze the deeper connotation of NMS from the perspective of graph theory, and then propose our optimization methods.

\section{A Graph Theory Perspective}\label{sec:graph}

The bottleneck of NMS algorithms lies in the extensive computation of IOUs. In practice, many IOUs are smaller than a given threshold $N_t$ and will not have any suppressive effect. We aim to consider only those IOUs that will affect the final result, thereby reducing the number of computations and improving efficiency. An IOU greater than $N_t$ indicates that two boxes have a suppressive effect on each other; otherwise, they are independent. We can treat this relationship as an edge in a graph, with each box as a node. This graph reflects the intrinsic structure of NMS, representing the connections between all boxes. By this transformation, we can directly analyze the NMS algorithm through the graph. Compared to a set of boxes in a two-dimensional plane, the structure of the graph is clearer and has more properties that can be utilized.

Specifically, we can regard the input $\mathcal{B},\mathcal{S},N_t$ of NMS algorithms as a directed graph $\mathcal{G}=(\mathcal{V},\mathcal{E})$. That's because we can think of every box in $\mathcal{B}$ as a node in a graph and draw an arc from $v$ to $u$ if box $v$ can suppress box $u$. Here, we give a formal definition.
\begin{definition}\label{def:graph:1}
Given a 3-tuple $(\mathcal{B}, \mathcal{S}, N_t)$ consisting of the bounding boxes, confidence scores and an IOU threshold, a graph $\mathcal{G}=(\mathcal{V},\mathcal{E})$ induced by NMS described as follows, there is an injective mapping of $\mathcal{B}$ into $\mathcal{V}$ that maps each bounding box $b_v$ in $\mathcal{B}$ to a node $v\in\mathcal{V}$, and for any ordered pair $(v,u)$, 
\[
\text{arc } (v, u)\in\mathcal{E}\iff s_v>s_u\wedge\text{IOU}(b_v, b_u)>N_t.
\]
\end{definition}
\begin{proposition}\label{pro:methodology:1}
$\mathcal{G}$ is a directed acyclic graph (DAG). 
\end{proposition}
We prove Proposition~\ref{pro:methodology:1} in the Appendix. Since $\mathcal{G}$ is a DAG, we can use dynamic programming to get the answer to NMS, i.e., $K$. Specifically, let $\mathit{\delta}(v)$ be the result of node $v$, i.e., $v$ is retained if $\mathit{\delta}(v)=1$, otherwise it's not. In original NMS, if there is a node $v$ that can suppress the current node $u$, then $u$ will not be retained. Therefore, we have the dynamic programming equation as follows, 
\[
\mathit{\delta}(u)=
\begin{cases}
\lnot\left(\bigvee_{v, (v,u)\in\mathcal{E}}\mathit{\delta}(v)\right) &\text{if }d^{-}(u)>0;\\
1 &\text{otherwise},
\end{cases}
\]
where $d^{-}(u)$ denotes the in-degree of $u$.

\begin{theorem}\label{thm:methodology:1}
$\forall k_i = K_{origin}[i]$, we have 
\[
k_i=\mathit{\delta}(i).
\]
\end{theorem}
Theorem~\ref{thm:methodology:1} shows that we can actually obtain the result through dynamic programming in topological sorting, shown in Figure~\ref{fig:methodology:1}. Because the result of DP depends only on valid topological sorts, which indicates that we do not need to sort confidence scores in descending order like original NMS to get the same answer, as long as the topological sort is valid. Additionally, we can observe that if there is no path from node $v$ to node $u$, then $v$ does not influence $u$. From this, we derive Corollary~\ref{cor:methodology:1}.
\begin{corollary}\label{cor:methodology:1}
If $v$ and $u$ are in two different weakly connected components (WCCs) of $\mathcal{G}$, then $\delta(v)$ and $\delta(u)$ are independent.
\end{corollary}

\begin{figure}[ht]
    \centering
    \subfigure[]{\label{subfig:1a}
        \begin{minipage}{.48\linewidth}
        \includegraphics[width=\linewidth]{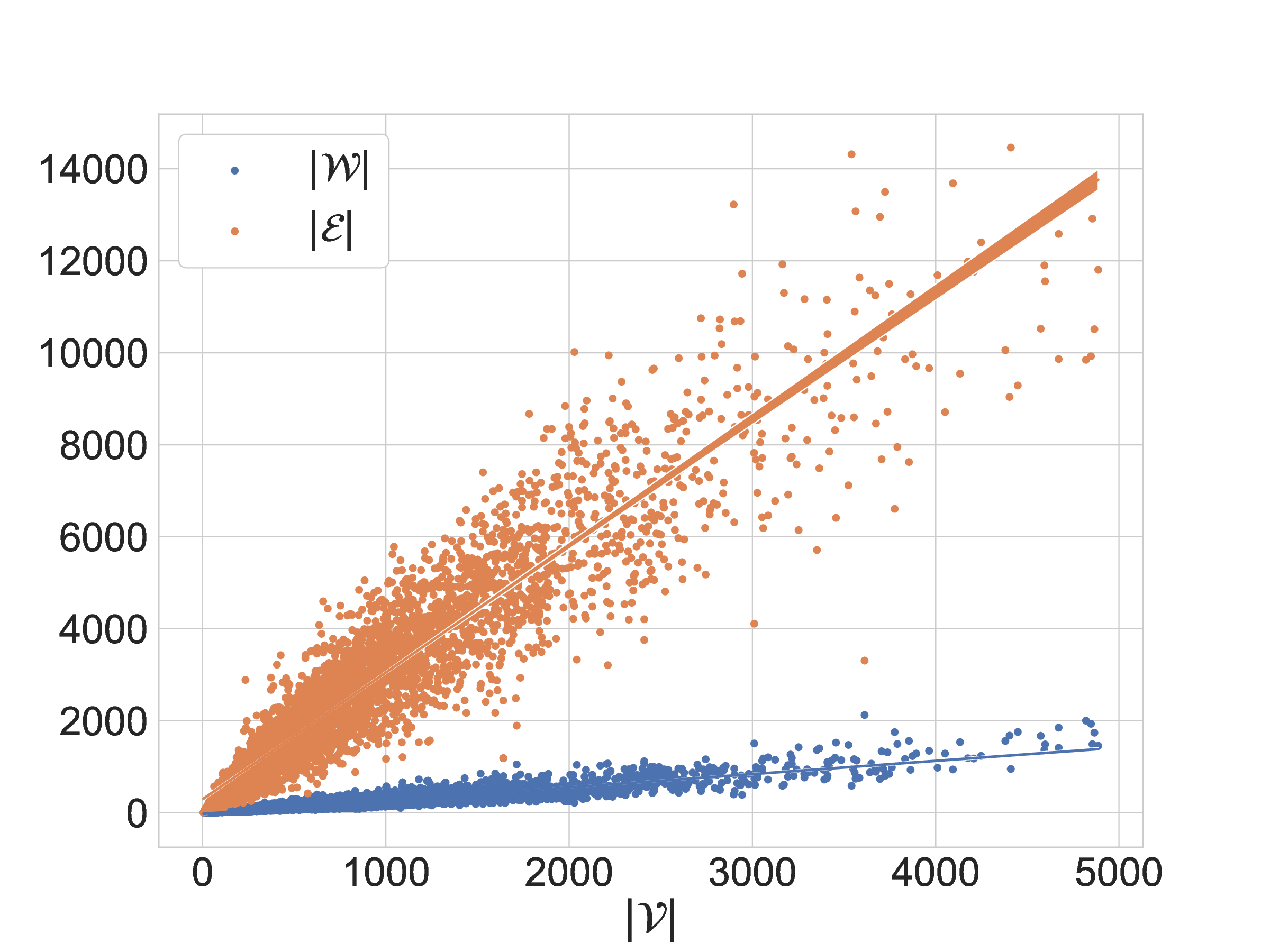}
        \end{minipage}
    }
    \subfigure[]{\label{subfig:1b}
        \begin{minipage}{.48\linewidth}
        \includegraphics[width=\linewidth]{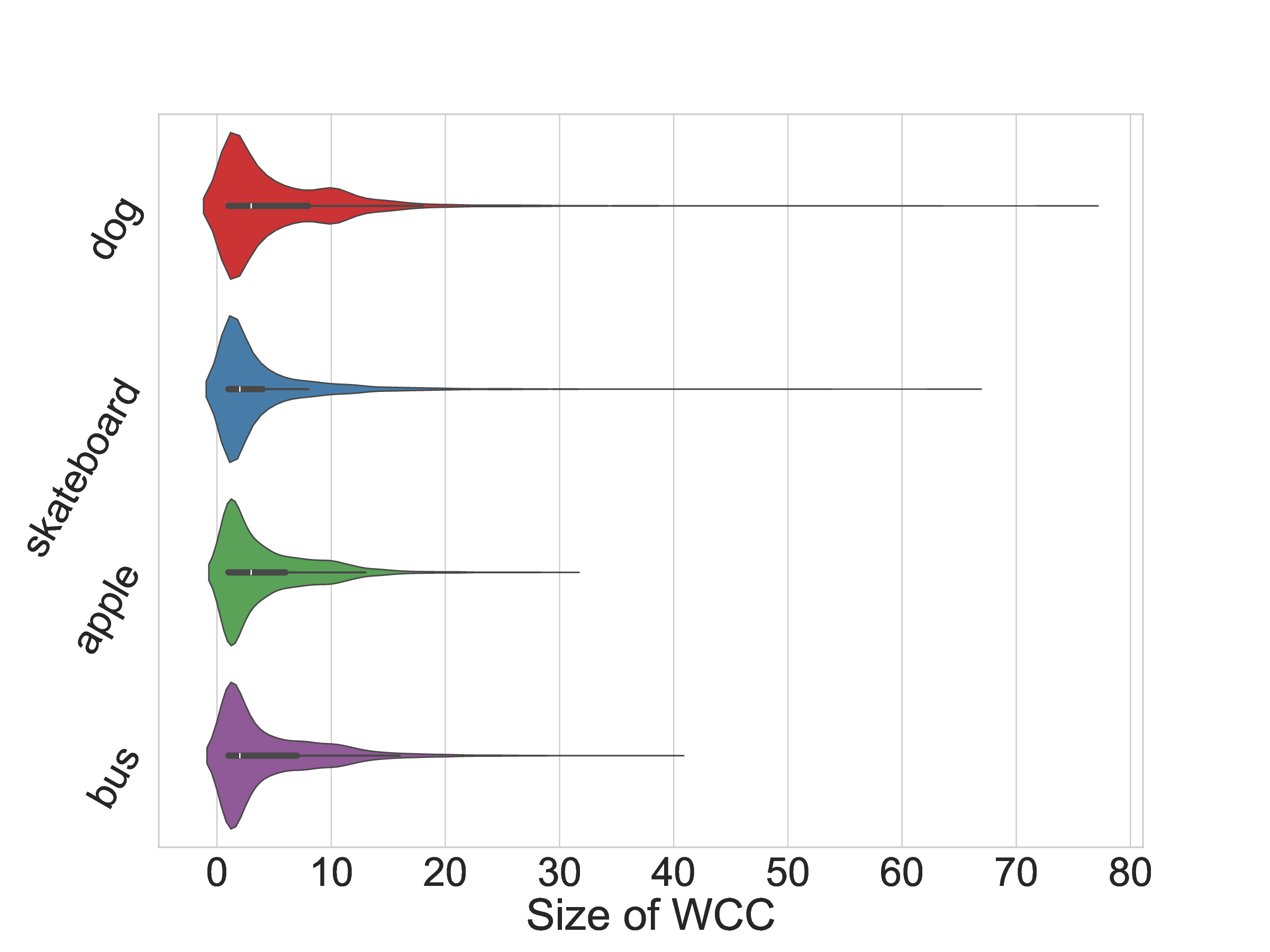}
        \end{minipage}
    }
    \caption{Statistical characteristics of graph $\mathcal{G}$ on MS COCO 2017 validation. \ref{subfig:1a} The scatter plot of $5000$ $\mathcal{G}$s on MS COCO 2017. It indicates that the number of arcs $\vert \mathcal{E}\vert$ and the number of WCCs $\vert\mathcal{W}\vert$ exhibit an approximately linear relationship with the number of nodes $\vert\mathcal{V}\vert$, respectively. \ref{subfig:1b} The violin plot of the sizes of WCCs across different categories on MS COCO 2017. It reveals the distributional characteristics of the sizes of the WCCs. It shows that over $50\%$ of the WCCs have a size less than $5$, and more than $75\%$ have a size less than $10$.}
    \label{fig:methodoloy:2}
\end{figure}

We find that completing dynamic programming in topological sorting requires $\mathcal{O}(\vert \mathcal{V}\vert + \vert \mathcal{E}\vert)$ time. In real-world data, $\vert\mathcal{E}\vert$ appears to have a linear relationship to $\vert\mathcal{V}\vert$ (see Figure~\ref{subfig:1a}), so once $\mathcal{G}$ is determined, NMS can be highly efficient via DP. However, quickly determining $\mathcal{G}$ is not a simple task. This is because, given a bounding box $b$, it is difficult to quickly determine which boxes in $\mathcal{B}$ have an $\text{IOU} > N_t$ with it. A related problem is improving the efficiency of the k-nearest neighbors algorithm (kNN), where \cite{kollar2006fast, elkin2023new,izbicki2015faster,beygelzimer2006cover} have made significant progress. However, IOU is more complex than the distance defined by norms, and we can only approximate $\mathcal{G}$ through related algorithms. We tried the latest research \cite{elkin2023new}, but it provided little help in acceleration due to its large constant.

\begin{figure}[h]
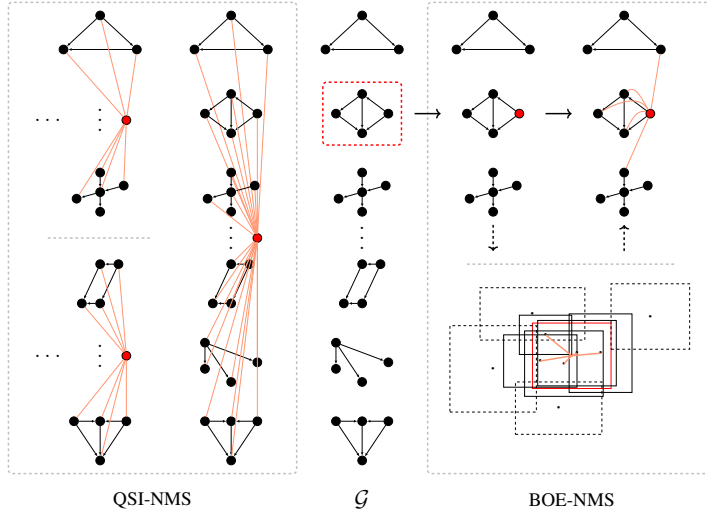

    \centering
    \include{tikz/mainidea}
    \caption{The key ideas behind QSI-NMS (left) and BOE-NMS (right). $\mathcal{G}$ (middle) contains many small weakly connected components (WCCs). QSI-NMS considers the global structure of the graph $\mathcal{G}$, where there are many WCCs. It selects a pivot (the red node on the left) and computes IOUs (orange edges) with all current subproblem nodes using a divide-and-conquer algorithm. BOE-NMS focuses on the local structure (the red dashed box) of $\mathcal{G}$, where most WCCs are quite small in size. It selects a node (the red node on the right) and only computes IOUs (orange edges) with its nearby nodes (solid arrows), which is derived from 2D plane geometric analysis (dashed arrows).}
    \label{fig:methodoloy:3}
\end{figure}

Fortunately, The NMS task is quite special, as its input comes from well-trained models, meaning that bounding boxes will cluster around many possible object locations, and bounding boxes predicted as different objects are independent of each other. This implies that $\mathcal{G}$ is a sparse graph with many WCCs, as shown in Figure~\ref{subfig:1a}. Additionally, we find that most of the WCCs are quite small, as shown in Figure~\ref{subfig:1b}. These two observations respectively suggest two optimization strategies (see Figure~\ref{fig:methodoloy:3}). Firstly, because WCCs are independent of each other, we can use a divide-and-conquer algorithm to break down many WCCs into fewer WCCs, continuously reducing the problem size to improve computational efficiency. Thus, we design QSI-NMS. Secondly, because most WCCs are quite small in size, we can reduce the cost of constructing arcs by geometric knowledge, leading to the design of BOE-NMS.

\section{Methodology}\label{methodology}

Following the graph-theoretic analysis of NMS in Section~\ref{sec:graph}, we propose two optimization methods based on two distinct characteristics of graph $\mathcal{G}$. Our approach is to design algorithms through the analysis of these characteristics to quickly construct $\mathcal{G}$ or an approximate graph $\tilde{\mathcal{G}}=(\tilde{\mathcal{V}},\tilde{\mathcal{E}})\approx \mathcal{G}$, enabling the use of dynamic programming in topological sorting to obtain NMS results. 

% Specifically, in Subsection~\ref{sec:qsi}, considering the high number of WCCs in $\mathcal{G}$, we introduce QSI-NMS. In Subsection~\ref{sec:boe}, noting that most of the WCCs are very small, we propose BOE-NMS.

\subsection{QSI-NMS}\label{sec:qsi}

We observe that graph $\mathcal{G}$ contains many WCCs, and according to Corollary~\ref{cor:methodology:1}, these components do not affect each other. This implies that, unlike the original NMS, which processes bounding boxes sequentially after sorting by confidence scores and is therefore very slow, we can solve the problem more efficiently using a divide-and-conquer algorithm, breaking it down into independent subproblems that can be solved recursively. Inspired by quicksort, we design quicksort induced NMS (QSI-NMS).

In each subproblem on $\mathcal{B}$, we can similarly select a pivot and calculate IOUs between the pivot and all the other boxes in $\mathcal{B}$, thereby constructing some arcs in $\mathcal{G}$. Next, we devise a partitioning criterion to split $\mathcal{B} \setminus \{\text{pivot}\}$ into two disjoint sets, $\mathcal{B}_l$ and $\mathcal{B}_r$, which are then solved recursively. Since IOUs are not calculated between boxes in $\mathcal{B}_l$ and $\mathcal{B}_r$, some arcs in the original $\mathcal{G}$ might be missed. Therefore, we need to carefully choose the pivot and partitioning criterion to ensure that the constructed $\tilde{\mathcal{G}}$ is as similar to $\mathcal{G}$ as possible.

For the pivot selection, we need to define a priority to choose the best pivot in $\mathcal{B}$. We find that selecting nodes with an in-degree of $0$ in $\mathcal{G}$ is optimal for two main reasons. First, node $v_0$ with an in-degree of $0$ belongs to some WCCs, and since most nodes in a WCC predict the same object, $v_0$ with the maximum confidence score will suppress most nodes, meaning it has many outgoing arcs. Choosing other nodes in the WCC might allocate $v_0$'s successors to different subsets, leading to significant discrepancies between $\mathcal{G}$ and $\mathcal{G}'$. Second, according to De Morgan's laws, the value of $\delta(u)$ is essentially the conjunction of the negations of the predecessors' $\delta$ values, formally described as follows:
\[ 
\delta(u) = \bigwedge_{v,(v,u)\in\mathcal{E}} \lnot \delta(v)
\]
This implies that missing an arc $(v_0, v)$ could result in $\delta(v)$ being incorrectly computed as $1$, causing a chain reaction that significantly deviates $K$ from ${K}_{origin}$. According to Definition~\ref{def:graph:1}, the node $v^*\in\mathcal{V}$ corresponding to the box $b^*\in\mathcal{B}$ with the highest confidence score has an in-degree of $0$. Hence, we select the box $b^*$ with the highest confidence score as the pivot.

For the partitioning criterion, we need to consider the spatial characteristics of different WCCs. Different WCCs are relatively dispersed in 2D space, so we can define the partitioning criterion based on the positions of the boxes. We represent the position of a box by its centroid, as it is intuitive and representative. Since the centroid is an ordered pair $(x,y)$, we can not compare it directly like real numbers. We need to define a preorder in $\mathbb{R}^2$. We find that different preorders have negligible effects on mAP. See Appendix \ref{discussion:qsi-eqsi} for details. We finally adopt the Manhattan distance to the origin $O(0, 0)$, i.e., the $L^1$ norm, as the comparison standard. Formally, we define a homogeneous relation $\preceq_{M}$ on $\mathbb{R}^2$:
\[ 
(x_1,y_1) \preceq_{M} (x_2,y_2) \Leftrightarrow \vert x_1\vert + \vert y_1\vert \leq \vert x_2\vert + \vert y_2\vert.
\]
Finally, we partition the set $\mathcal{B}\setminus \{b^*\}$ as follows:
\[
\begin{cases}
    \mathcal{B}_l = \{b_c\vert b_c\preceq_{M} b^*_c\wedge b\in\mathcal{B}\setminus\{b^*\}\};\\
    \mathcal{B}_r = \{b_c\vert b_c\npreceq_{M} b^*_c\wedge b\in\mathcal{B}\setminus\{b^*\}\},
\end{cases}
\]
where $b_c$ and $b^*_c$ denote the centroid of $b$ and $b^*$, respectively. Since we always choose the box with the highest confidence score, this creates a valid topological sort of $\tilde{\mathcal{G}}$. Thus, we can avoid explicitly constructing $\tilde{\mathcal{G}}$, further reducing computational overhead. The pseudo-code for QSI-NMS can be found in the Appendix.

\paragraph{eQSI-NMS Taking $\mathcal{O}(n\log n)$ Time} Though QSI-NMS performs very well in the real world, it is not an $\mathcal{O}(n\log n)$ algorithm for the simple reason that the pivot is not chosen randomly. By analyzing the structure of QSI-NMS, we further optimize it and propose extended QSI-NMS (eQSI-NMS), which only takes $\mathcal{O}(n \log n)$ time. Since in carrying out QSI-NMS we always split the problem into two subproblems, we can thus construct a binary tree.
\begin{definition}\label{def:methodology:2}
Given a 3-tuple $(\mathcal{B}, \mathcal{S}, N_t)$, a QSI-tree for $\mathcal{B}$ denoted by $QT(\mathcal{B})$ is a binary tree defined recursively as follow:
\begin{itemize}
    \item Its root is a node corresponding to the box $b_v\in\mathcal{B}$ with maximum confidence $s_v$.
    \item Its left subtree is $QT(\mathcal{B}_{l})$, where $\mathcal{B}_l$ is the left subset of $\mathcal{B}$ in QSI-NMS.
    \item Its right subtree is  $QT(\mathcal{B}_r)$, where $\mathcal{B}_r$ is the right subset of $\mathcal{B}$ in QSI-NMS. 
\end{itemize}
The basic case is that if $\mathcal{B}$ is empty, then QSI-tree is also empty, i.e., $QT(\emptyset)=\emptyset$.
\end{definition}

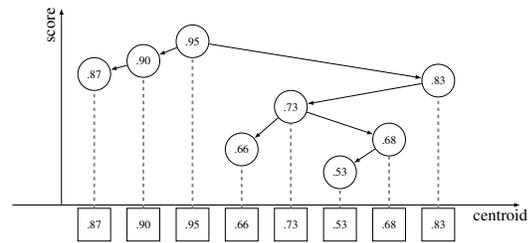
\begin{wrapfigure}{r}{.5\linewidth}
    \centering
    % \documentclass[12pt,a4paper]{article}
% \RequirePackage[left=2.50cm, right=2.50cm, top=2.50cm, bottom=2.50cm]{geometry}
% \usepackage[fontsize=12pt]{fontsize}

% \usepackage{
%     algorithm,
%     algorithmicx,
%     algpseudocode,
%     authblk, 
%     graphicx, 
%     amsthm, 
%     amsmath, 
%     amssymb, 
%     booktabs, 
%     listings, 
%     xcolor,
%     url,
%     tcolorbox,
%     fancyhdr
% }

% \usepackage{tikz}
% \usetikzlibrary{automata, positioning, arrows.meta, shapes, backgrounds, datavisualization,calc}

% % +--------------------------------------------+
% % |                  /\_/\                     |
% % |                 ( o.o )                    | 
% % |                  > ^ <                     |
% % |            have fun writing                |
% % +--------------------------------------------+

% \begin{document}

\resizebox{\linewidth}{!}{
\begin{tikzpicture}

\tikzset{
    block/.style={
        minimum width=1cm, minimum height=1cm, thick, draw
    },
    uv/.style={
        minimum size=1cm, thick, draw, circle
    },
    arc/.style={
        -latex, thick
    },
    al/.style={
        draw=gray, dashed, ultra thick
    },
}

% \draw [help lines] (0, 0) grid (9, 6);

\draw [-latex, thick] (-1.5, 1) -- (14, 1);
\draw [-latex, thick] (0, 1) -- (0, 7);
% \draw [thick] (0, 7) -- (14, 7);
% \draw [thick] (14, 7) -- (14, 1);

\node [] at (13.4, 0.69) {\Large centroid};
\node [rotate=90] at (-.31, 6.5) {\Large score};

\node (11) [block] at (1, 0.4) {.87};
\node (22) [block] at (2.5, 0.4) {.90};
\node (33) [block] at (4, 0.4) {.95};
\node (44) [block] at (5.5, 0.4) {.66};
\node (55) [block] at (7, 0.4) {.73};
\node (66) [block] at (8.5, 0.4) {.53};
\node (77) [block] at (10, 0.4) {.68};
\node (88) [block] at (11.5, 0.4) {.83};

\node (1) [uv] at (1, 5) {.87};
\node (2) [uv] at (2.5, 5.4) {.90};
\node (3) [uv] at (4, 6) {.95};
\node (4) [uv] at (5.5, 2.7) {.66};
\node (5) [uv] at (7, 4) {.73};
\node (6) [uv] at (8.5, 2) {.53};
\node (7) [uv] at (10, 3) {.68};
\node (8) [uv] at (11.5, 4.8) {.83};

\draw [arc] (2) -- (1);
\draw [arc] (3) -- (2);
\draw [arc] (3) -- (8);
\draw [arc] (8) -- (5);
\draw [arc] (5) -- (4);
\draw [arc] (5) -- (7);
\draw [arc] (7) -- (6);

\foreach \v in {1, 2, ..., 8} {
    \draw [al] (\v\v) -- (\v);
}

% \coordinate (O) at (0,0);
% \coordinate (1) at ($(O) + (9.4, 18.8)$);
% \coordinate (2) at ($(O) + (4.8, 15.4)$);
% \coordinate (3) at ($(O) + (15, 10)$);
% \coordinate (4) at ($(O) + (1.1, 5.2)$);
% \coordinate (5) at ($(O) + (8.2, 2.4)$);
% \coordinate (6) at ($(O) + (12.5, 7.1)$);
% \coordinate (7) at ($(O) + (17.5, 6.8)$);

% \coordinate[label=left:\large Scores] (X_top) at ($(O) + (0, 20)$);
% \coordinate[label=below:\large $\leq_{(x,y)}$](Y_top) at ($(O) + (20, 0)$);

% % XY
% \draw[very thick, arrows = {-Computer Modern Rightarrow[line join=miter]}] (O) -- (X_top);
% \draw[very thick, arrows = {-Computer Modern Rightarrow[line join=miter]}] (O) -- (Y_top);

% \foreach \i/\texti  in {1,2,3,4,5,6,7} {
%     \node[draw, shape = circle, line width=0.8pt,inner sep=3pt, fill=black] at (\i) (node\i) {};
% }

% % \draw[ thick, ->] (1) -- ($(1) + (-9, 0)$);
% % \draw[ thick, ->] (1) -- ($(1) + (9.3, 0)$);

% % \draw[ thick, ->] (2) -- ($(2) + (-4, 0)$);
% % \draw[ thick, ->] (2) -- ($(2) + (4.3, 0)$);

% % \draw[ thick, ->] (3) -- ($(3) + (-5, 0)$);
% % \draw[ thick, ->] (3) -- ($(3) + (4, 0)$);

% \draw[densely dashed, thick]  ($(1) + (0, 3.2)$) -- ($(1) + (0, -20)$);
% \draw[densely dashed, thick]  ($(2) + (0, 3.2)$) -- ($(2) + (0, -16.7)$);
% \draw[densely dashed, thick]  ($(3) + (0, 8)$) -- ($(3) + (0, -11.3)$);

% \draw[thick, ->] (node1) -- (node2);
% \draw[thick, ->] (node1) -- (node3);
% \draw[thick, ->] (node2) -- (node4);
% \draw[thick, ->] (node2) -- (node5);
% \draw[thick, ->] (node3) -- (node6);
% \draw[thick, ->] (node3) -- (node7);

\end{tikzpicture}

}
% \end{document}
    \caption{A Cartesian tree for $B$. The x-axis represents the centroid, where the node on the left $\preceq_{\mathcal{C}}$ the one on the right. The y-axis represents the confidence score, where the node below $\preceq_{\mathcal{P}}$ the one above. The values of the sequence below the x-axis are the confidence scores of $B$.}
    \label{fig:methodology:2}
\end{wrapfigure}

An example of QSI-tree is shown in Figure~\ref{fig:methodology:2}. QSI-tree reveals the inherent structure of QSI-NMS, allowing us to consider QSI-NMS from a high-level perspective. More generally, in QSI-NMS, we tag each box with an ordered pair $(p, c)$, where $p \in \mathcal{P}$ represents the priority and $c \in \mathcal{C}$ is the key used for partitioning. We define preorder relations $\preceq_{\mathcal{P}}$ on $\mathcal{P}$ and $\preceq_{\mathcal{C}}$ on $\mathcal{C}$. This indicates that the QSI-tree is essentially a binary search tree that satisfies the max-heap property: the priority of the parent node is not less than that of the child nodes, the keys in the left subtree are all less than or equal to the parent node, and the keys in the right subtree are all greater than the parent node. Furthermore, we have the following Theorem~\ref{thm:methodology:2}.

\begin{theorem}\label{thm:methodology:2}
We sort all the elements of $\mathcal{B}$ in ascending order of boxes' centroids according to the preorder $\preceq_{\mathcal{C}}$ into a sequence:
\[
B=(b_{i_1},b_{i_2},\ldots,b_{i_n}),
\]
then QSI-tree is a Cartesian tree for $B$ in which each key is the confidence score of the corresponding box. 
\end{theorem} 
According to the dynamic programming, if a node $v$ can affect the result $\delta(u)$ of node $u$, there must exist a path between $v$ and $u$. In QSI-NMS, this manifests as node $v$ only being able to influence nodes within its subtrees in QSI-tree. Theorem~\ref{thm:methodology:2} states that QSI-tree is a Cartesian tree, indicating that the subtree of $v$ corresponds to a contiguous interval in $B$, as shown in Figure~\ref{fig:methodology:2}.

% Theorem~\ref{thm:methodology:2} shows a QSI-tree is a Cartesian tree, which implies that the nodes of $v$'s subtree fall in a continuous interval $B[l_v,r_v]$ of the sequence $B$, where $l_v$ is the last position among the previous positions that contain a larger value than $s_v$ while $r_v$ is the first position among the following positions that contain a larger value than $s_v$. 

Specifically, the subtree of $v$ corresponds to the contiguous interval $B[l_v+1:r_v-1]$ in $B$, where $l_v$ is the last position before $v$ that is greater than $s_v$, and $r_v$ is the first position after $v$ that is greater than $s_v$. Finding $(l_v, r_v)$ for all $v$ is known as the all nearest greater values problem, which can be solved in $\mathcal{O}(n)$ time by maintaining a stack. Similarly to QSI-NMS, we can complete the suppression during the algorithm. Therefore, we obtain the following time complexity:
\[
\mathcal{O}(n\log n+n)=\mathcal{O}(n\log n).
\]
According to our best knowledge, this algorithm is the most optimal in terms of complexity. The pseudo-code can also be found in the Appendix.

% The solution to find $(l_v,r_v)$ for each node $v$ is similar to the all nearest smaller values problem, and the only difference is that we're seeking the maximum value instead of the minimum. After we sort the boxes to get $B$, we can use a stack-based algorithm taking $\mathcal{O}(n)$ time to solve this problem. 

\subsection{BOE-NMS}\label{sec:boe}

% In QSI-NMS, we can explicitly or implicitly construct the $\tilde{\mathcal{G}}$ so as to derive the results of NMS by dynamic programming. For efficiency, we construct $\tilde{\mathcal{G}}$ recursively. As a result, it causes some arcs that would otherwise be in $\mathcal{G}$ to be ignored, which makes QSI-NMS a negligible loss in $e(\mathcal{K})$, e.g., a loss of $0.1\%$ mAP in YOLOv8-N model on MS COCO. Formally, $\tilde{\mathcal{G}}=(\tilde{\mathcal{V}},\tilde{\mathcal{E}})$ and $\mathcal{G}=(\mathcal{V}, \mathcal{E})$ are related as follows,
% \[
% \begin{cases}
%     \tilde{\mathcal{V}}=\mathcal{V},\\
%     \tilde{\mathcal{E}}\subset \mathcal{E}.
% \end{cases}
% \]
% We focus on the locality of bounding box locations, and we find that a box can only have a large IOU with its neighbors, which implies that $\mathcal{G}$ is a sparse graph. We have the following formal theorem at our disposal.
We find that the vast majority of WCCs in $\mathcal{G}$ are very small, as shown in Figure~\ref{subfig:1b}. This is because there are not many bounding boxes predicting the same object, and NMS is a form of local suppression. We hope to consider the locality of box distributions, so that the currently selected box only computes IOUs with boxes corresponding to nodes in the same WCC, rather than computing IOUs with all boxes as in original NMS.

We focus on the spatial locality of boxes. We found that a box is likely to have large IOUs only with neighbors that are relatively close to it in 2D space, which also indicates that $\mathcal{G}$ is a sparse graph. Formally, we have the following theorem:
\begin{theorem}\label{thm:methodology:3}
    Given a bounding box $b^{*}\in\mathcal{B}$, $\forall b\in\mathcal{B}$, we have $\text{IOU}(b^{*},b)\leq\frac{1}{2}$ if the centroid of $b$ does not lie within $b^{*}$. Formally,
    \[
    % \begin{cases}
    %     x^{(b)}_{c}>x^{(b^{*})}_{rb}\vee x^{(b)}_{c}<x^{(b^{*})}_{lt};\\
    %     y^{(b)}_{c}>y^{(b^{*})}_{rb}\vee y^{(b)}_{c}<y^{(b^{*})}_{lt},
    % \end{cases}
    \left(x^{(b)}_{c}>x^{(b^{*})}_{rb}\vee x^{(b)}_{c}<x^{(b^{*})}_{lt}\right)\vee \left(y^{(b)}_{c}>y^{(b^{*})}_{rb}\vee y^{(b)}_{c}<y^{(b^{*})}_{lt}\right),
    \]
    where $(x^{(b)}_c,y^{(b)}_c)$, $(x^{(b^{*})}_{lt}, y^{(b^{*})}_{lt})$ and $(x^{(b^{*})}_{rb},y^{(b^{*})}_{rb})$ denote the coordinates of the centroid of $b$, the left-top and the right-bottom corners of $b^*$, respectively.
\end{theorem}
Since $N_t$ is usually greater than $0.5$, e.g., $0.7$ for YOLOv8 and Faster R-CNN. By Theorem~\ref{thm:methodology:3} we can check IOUs only for those boxes whose centorids lie within the current box. Based on this, we propose boxes outside excluded NMS (BOE-NMS), a method devoid of mAP loss.

In BOE-NMS, We first sort the boxes by their centroids according to lexicographic order $\preceq_{L}$ on $\mathbb{R}^2$ which is defined as follows:
\[
(x_1,y_1)\preceq_{L}(x_2,y_2)\Leftrightarrow (x_1<x_2)\vee (x_1=x_2\wedge y_1\leq y_2).
\]
Then for the current box $b^*$, we can find all the boxes whose centroids may lie in $b^*$ in $\mathcal{O}(\log n)$ time, and we just need to check one by one whether the IOUs between $b$ and these boxes are greater than $N_t$. The pseudo-code for BOE-NMS is described in Algorithm~\ref{alg:boe-nms} which can be found in the Appendix. Let's set aside Theorem~\ref{thm:methodology:3} for now. A more intuitive but weaker conclusion is that if two boxes do not intersect, their IOU must be $0$. However, this is not conducive to efficient implementation because of the high cost of maintaining the corresponding data structure. We discuss this issue in the Appendix.

$N_t$ is typically set to $0.7$, and the method based on Theorem~\ref{thm:methodology:3} does not introduce errors. We also provide a tighter bound to further optimize BOE-NMS. Based on Theorem~\ref{thm:methodology:4} which is a generalization of Theorem~\ref{thm:methodology:3}, we can handle cases where $N_t$ is any real number $\in(0,1)$.
\begin{theorem}\label{thm:methodology:4}
We use $s$ to denote a scaling factor, and then we can use $\alpha(b, s)$ to represent the new box $b'$ obtained by scaling $b$. Formally, 
\[
\begin{cases}
    x^{(b')}_{lt} = x^{(b)}_{c} - s\times \vert x^{(b)}_{lt}-x^{(b)}_{c}\vert,\\
    x^{(b')}_{rb} = x^{(b)}_{c} + s\times \vert x^{(b)}_{rb}-x^{(b)}_{c}\vert,\\
    y^{(b')}_{lt} = y^{(b)}_{c} - s\times \vert y^{(b)}_{lt}-y^{(b)}_{c}\vert,\\
    y^{(b')}_{rb} = y^{(b)}_{c} + s\times \vert y^{(b)}_{rb}-y^{(b)}_{c}\vert.
\end{cases}
\]
Given any $N_t\in(0,1)$, if the centroid of $b$ does not lie within $\alpha(b^*,\nicefrac{1}{N_t}-1)$, then $\text{IOU}(b,b^*)\leq N_t$.
\end{theorem}

Since BOE-NMS only excludes boxes with IOU $\leq N_t$, the graph constructed by the BOE-NMS is the same as $\mathcal{G}$. In other words, the results of BOE-NMS are identical to original NMS. However, unlike original NMS, BOE-NMS does not need to compute IOUs with all remaining boxes but rather determines the boxes that could potentially be suppressed in $\mathcal{O}(\log n)$ time. Next, inspect each of these $t$ ($t\approx \text{size of the corresponding WCC}$) boxes one by one in $\mathcal{O}(t)$ time. As shown in Figure~\ref{subfig:1b}, the sizes of weakly connected components are almost all less than a constant, say $10$. This means that the actual performance of BOE-NMS approaches linear time complexity, but strictly speaking, the complexity is still $\mathcal{O}(n^2)$.

% Unlike original NMS, in each iteration, BOE-NMS does not need to compute the IOUs with all remaining boxes, but instead determines those boxes where the IOUs are likely to be greater than the threshold in $\mathcal{O}(\log n)$ time. 
% As a result, BOE-NMS optimizes original NMS at the constant level and is $5x$(TODO) faster than original NMS, but still runs in $\mathcal{O}(n^2)$ time.

\section{Experiments}\label{exp}

In this section, we first introduce NMS-Bench, the first end-to-end benchmark for rapid validation of NMS algorithms. Next, we validate our algorithms on NMS-Bench and compare them with classical algorithms: original NMS \cite{greedynms_human_dectection}, Fast NMS \cite{yoloact}, and Cluster-NMS \cite{ClusterNMS}. We conduct tests on MS COCO 2017 \cite{ms-coco} and Open Images V7 \cite{OpenImages} using YOLOv5 \cite{yolov5}, YOLOv8 \cite{yolov8}, and Faster R-CNN \cite{fastrcnn} as validation models. More experimental details can be found in the Appendix.

\subsection{NMS-Bench}

NMS-Bench is a robust framework that allows researchers to evaluate various NMS methods over different models and datasets in a few minutes. NMS-Bench primarily consists of three components: original bounding box data without NMS applied, implementations of various NMS algorithms as benchmarking methods, and evaluation metrics. The code for NMS-Bench is available on GitHub\footnote{https://github.com/Yuri3-xr/NMS-Bench}.

For the original boxes, we extracted non-NMS boxes using different models (YOLO series \cite{YOLO, yolov8} and Faster R-CNN \cite{fastrcnn}) on various datasets \cite{ms-coco, OpenImages} to create the NMS-Bench dataset, thereby decoupling the model inference and post-processing stages. This approach saves significant computational resources during inference. We provide a large amount of data for testing, including original boxes from a total of 273,100 images. More detailed information can be found in the Appendix.

For benchmarking methods, NMS-Bench implements classical algorithms such as original NMS \cite{greedynms_human_dectection}, Fast NMS \cite{yoloact}, Cluster-NMS \cite{ClusterNMS}, and PSRR-MaxpoolNMS \cite{Zhang_2021_CVPR}. QSI-NMS (including eQSI-NMS) and BOE-NMS are also included in NMS-Bench. These methods enable researchers to reproduce and study NMS algorithms. All algorithms are implemented fairly. Researchers can also quickly implement and validate their own NMS algorithms, as NMS-Bench is a plug-and-play, end-to-end benchmark.

For evaluation metrics, we use COCO-style mAP as the accuracy metric and average processing latency per image as the efficiency metric. The latency calculation begins from the input of bounding boxes and ends when the retained bounding boxes are output. For a dataset containing $N$ images, latency is measured by using the bounding boxes generated per image as input, and the total latency for the $N$ images is averaged. To mitigate random errors, this measurement is repeated 5 times, and the average of these measurements is used as the final average latency. 

% The source code can be found on our github repository: \href{}{TODO}.

% \paragraph{Raw Data} To alleviate the computational cost of validating NMS, we extracted raw bounding boxes generated from different models (YOLO series \cite{YOLO,yolov2,yolov3} and Faster R-CNN \cite{FasterRCNN}) across different datasets (MS COCO \cite{ms-coco} and Open Images \cite{OpenImages}) as the raw data.

% \paragraph{NMS Algorithms} In NMS-Bench, we have implemented QSI-NMS and BOE-NMS. Additionally, we have impartially implemented some existing NMS algorithms (such as original NMS, Fast NMS, Cluster-NMS, etc.) as benchmarks to facilitate experimentation for researchers. Other NMS methods can also be easily implemented and validated here, as we provide corresponding interfaces.

% \paragraph{Metrics} We selected the COCO-style mAP calculation method as our accuracy metric. For our performance metric, we used the average latency per image, which is derived by dividing the total computation time of the NMS algorithm on the dataset by the number of images.

\subsection{Results}

In Table~\ref{tab:COCO}, we compare our methods with some mainstream work on MS COCO 2017. We observe that our methods, particularly eQSI-NMS, demonstrate substantial performance enhancements in processing speed across different models on MS COCO 2017. eQSI-NMS stands out by offering up to $16.9 \times$ speed of  original NMS, $4.3 \times$ speed of Fast NMS, and $8.9 \times$ speed of Cluster-NMS with a mAP decrease of about $0.5\%$. Similarly, QSI-NMS provides $8.8 \times$ speed of original NMS, $2.2 \times$ speed of Fast NMS, and $4.6 \times$ speed of Cluster-NMS with a marginal mAP decrease of about $0.2\%$. BOE-NMS also shows significant enhancements, being $9.1 \times$ as fast as original NMS, $2.3 \times$ as fast as Fast NMS, and $4.8 \times$ as fast as Cluster-NMS with no mAP decrease.

Table~\ref{tab:OI} shows that on Open Images V7, eQSI-NMS provides approximately $10.2 \times$ speed of original NMS, $3.7 \times$ speed of Fast NMS, and $7.0 \times$ speed of Cluster-NMS. QSI-NMS is about $5.6 \times$ as fast as original NMS, $2.0 \times$ as fast as Fast NMS, and $3.9 \times$ as fast as Cluster-NMS. Similarly, BOE-NMS achieves $5.4 \times$ speed of original NMS, $2.0 \times$ speed of Fast NMS, and $3.8 \times$ speed of Cluster-NMS. On Open Images V7, QSI-NMS and eQSI-NMS perferm well in mAP with about $0.2\%$ mAP decreasing.

\begin{table}[h]
    \centering
    \caption{NMS Methods Performance on MS COCO 2017}
    \label{tab:COCO}
    \resizebox{\linewidth}{!}{
        \begin{tabular}{@{}ccccccccc@{}}
            \toprule
            \textbf{Model}                         & \textbf{Size}      & \textbf{Target}          & \textbf{original NMS} & \textbf{Fast NMS} & \textbf{Cluster-NMS} & \textbf{BOE-NMS} & \textbf{QSI-NMS} & \textbf{eQSI-NMS} \\
            \midrule
            \multirow{10}{*}{YOLOv8}               & \multirow{2}{*}{N} & Average Latency ($\mu$s) & 906.9                 & 321.4             & 600.8                & 176.8            & 146.8            & \textbf{85.0}     \\
                                                   &                    & AP 0.5:0.95 (\%)            & 37.2                  & 37.0              & 37.2                 & 37.2             & 37.1             & 36.9              \\
                                                   & \multirow{2}{*}{S} & Average Latency ($\mu$s) & 531.2                 & 232.5             & 421.5                & 126.1            & 109.4            & \textbf{63.4}     \\
                                                   &                    & AP 0.5:0.95 (\%)            & 44.8                  & 44.6              & 44.8                 & 44.8             & 44.6             & 44.5              \\
                                                   & \multirow{2}{*}{M} & Average Latency ($\mu$s) & 353.3                 & 202.6             & 348.5                & 100.8            & 93.1             & \textbf{53.7}     \\
                                                   &                    & AP 0.5:0.95 (\%)            & 50.2                  & 50.0              & 50.2                 & 50.2             & 50.0             & 49.9              \\
                                                   & \multirow{2}{*}{L} & Average Latency ($\mu$s) & 196.5                 & 51.3              & 90.7                 & 82.1             & 67.1             & \textbf{39.0}     \\
                                                   &                    & AP 0.5:0.95 (\%)            & 52.8                  & 52.6              & 52.8                 & 52.8             & 52.7             & 52.5              \\
                                                   & \multirow{2}{*}{X} & Average Latency ($\mu$s) & 183.0                 & 148.6             & 262.2                & 69.2             & 66.8             & \textbf{39.6}     \\
                                                   &                    & AP 0.5:0.95 (\%)            & 53.9                  & 53.7              & 53.9                 & 53.9             & 53.8             & 53.6              \\
            \midrule
            \multirow{10}{*}{YOLOv5}               & \multirow{2}{*}{N} & Average Latency ($\mu$s) & 10034.2               & 1724.2            & 3949.1               & 719.6            & 688.9            & \textbf{339.0}    \\
                                                   &                    & AP 0.5:0.95 (\%)            & 27.8                  & 27.6              & 27.8                 & 27.8             & 27.5             & 27.4              \\
                                                   & \multirow{2}{*}{S} & Average Latency ($\mu$s) & 4441.4                & 996.4             & 2152.5               & 438.1            & 449.2            & \textbf{226.5}    \\
                                                   &                    & AP 0.5:0.95 (\%)            & 37.2                  & 36.9              & 37.2                 & 37.2             & 36.9             & 36.6              \\
                                                   & \multirow{2}{*}{M} & Average Latency ($\mu$s) & 3351.6                & 874.1             & 1851.2               & 350.5            & 408.3            & \textbf{204.9}    \\
                                                   &                    & AP 0.5:0.95 (\%)            & 45.1                  & 44.8              & 45.1                 & 45.1             & 44.9             & 44.5              \\
                                                   & \multirow{2}{*}{L} & Average Latency ($\mu$s) & 2531.2                & 732.8             & 1484.2               & 286.3            & 353.3            & \textbf{178.4}    \\
                                                   &                    & AP 0.5:0.95 (\%)            & 48.8                  & 48.4              & 48.8                 & 48.8             & 48.6             & 48.2              \\
                                                   & \multirow{2}{*}{X} & Average Latency ($\mu$s) & 1959.1                & 630.8             & 1273.9               & 248.5            & 314.7            & \textbf{160.3}    \\
                                                   &                    & AP 0.5:0.95 (\%)            & 50.5                  & 50.1              & 50.5                 & 50.5             & 50.3             & 49.9              \\
            \midrule
            \multirow{2}{*}{Faster R-CNN R50-FPN}  & \multirow{2}{*}{-} & Average Latency ($\mu$s) & 57.2                  & 112.0             & 184.4                & 41.1             & 36.6             & \textbf{25.7}     \\
                                                   &                    & AP 0.5:0.95 (\%)            & 39.8                  & 39.9              & 39.8                 & 39.8             & 39.5             & 39.3              \\
            \multirow{2}{*}{Faster R-CNN R101-FPN} & \multirow{2}{*}{-} & Average Latency ($\mu$s) & 49.5                  & 100.2             & 175.8                & 37.1             & 33.9             & \textbf{23.9}     \\
                                                   &                    & AP 0.5:0.95 (\%)            & 41.8                  & 41.7              & 41.8                 & 41.8             & 41.5             & 41.4              \\
            \multirow{2}{*}{Faster R-CNN X101-FPN} & \multirow{2}{*}{-} & Average Latency ($\mu$s) & 39.7                  & 95.8              & 169.7                & 31.9             & 30.1             & \textbf{21.4}     \\
                                                   &                    & AP 0.5:0.95 (\%)            & 43.0                  & 42.8              & 43.0                 & 43.0             & 42.7             & 42.5              \\
            \bottomrule
        \end{tabular}
    }
\end{table}

\begin{table}[htbp]
    \centering
    \caption{NMS Methods Performance on Open Images V7}
    \label{tab:OI}
    \resizebox{\linewidth}{!}{
        \begin{tabular}{@{}cccccccccc@{}}
            \toprule
            \textbf{Model}           & \textbf{Size}      & \textbf{Target}          & \textbf{original NMS} & \textbf{Fast NMS} & \textbf{Cluster-NMS} & \textbf{BOE-NMS} & \textbf{QSI-NMS} & \textbf{eQSI-NMS} \\
            \midrule
            \multirow{10}{*}{YOLOv8} & \multirow{2}{*}{N} & Average Latency ($\mu$s) & 1627.9             & 498.1         & 952.4            & 260.2        & 231.8         & \textbf{132.3} \\
                                     &                    & AP 0.5:0.95 (\%)            & 18.1                  & 18.2              & 18.1                 & 18.1             & 18.1             & 18.0              \\
                                     & \multirow{2}{*}{S} & Average Latency ($\mu$s) & 1212.1             & 412.8         & 781.6            & 214.7         & 199.5         & \textbf{111.2} \\
                                     &                    & AP 0.5:0.95 (\%)            & 27.3                  & 27.2              & 27.3                 & 27.3             & 27.1             & 27.1              \\
                                     & \multirow{2}{*}{M} & Average Latency ($\mu$s) & 1003.3             & 371.6         & 744.0            & 191.9         & 189.0         & \textbf{100.6} \\
                                     &                    & AP 0.5:0.95 (\%)            & 33.1                  & 33.1              & 33.1                 & 33.1             & 33.0             & 32.9              \\
                                     & \multirow{2}{*}{L} & Average Latency ($\mu$s) & 853.1             & 350.5         & 665.6            & 175.0         & 180.3         & \textbf{100.8} \\
                                     &                    & AP 0.5:0.95 (\%)            & 34.4                  & 34.4              & 34.4                 & 34.4             & 34.3             & 34.2              \\
                                     & \multirow{2}{*}{X} & Average Latency ($\mu$s) & 803.4             & 342.0         & 648.8            & 168.5         & 173.2         & \textbf{95.1} \\
                                     &                    & AP 0.5:0.95 (\%)            & 35.9                  & 35.8              & 35.9                 & 35.9             & 35.8             & 35.7              \\
            \bottomrule
        \end{tabular}
    }
\end{table}

\begin{table}[htbp]
    \centering
    \caption{Comparisons of Our Methods and PSRR-MaxpoolNMS}
    \label{tab:psrr-anchor-base}
    \resizebox{\linewidth}{!}{
        \begin{tabular}{@{}ccccccccc@{}}
            \toprule
            \textbf{Model}           & \textbf{Size}      & \textbf{Target}          & \textbf{original NMS} & \textbf{PSRR-MaxpoolNMS} & \textbf{BOE-NMS} & \textbf{QSI-NMS} & \textbf{eQSI-NMS} \\
            \midrule
            \multirow{6}{*}{YOLOv5}
&\multirow{2}{*}{N}&Average Latency ($\mu$s) &8568.5&599.6&906.2&628.0&\textbf{325.6}\\
&& AP 0.5:0.95 (\%)&27.8&26.5&27.8&27.5&27.4\\
&\multirow{2}{*}{S}&Average Latency ($\mu$s) &3858.2&409.4&547.7&408.2&\textbf{217.5}\\
&& AP 0.5:0.95 (\%)&37.2&35.6&37.2&36.9&36.6\\
&\multirow{2}{*}{M}&Average Latency ($\mu$s) &2918.1&380.5&424.7&371.3&\textbf{197.4}\\
&& AP 0.5:0.95 (\%)&45.1&43.1&45.1&44.9&44.5\\
            \midrule
            \multirow{2}{*}{Faster R-CNN R50-FPN}
&\multirow{2}{*}{-}&Average Latency ($\mu$s) &53.0&89.0&43.1&34.3&\textbf{24.6}\\
&& AP 0.5:0.95 (\%)&39.8&37.5&39.8&39.5&39.3\\
            \multirow{2}{*}{Faster R-CNN R101-FPN}
&\multirow{2}{*}{-}&Average Latency ($\mu$s) &45.5&86.4&38.6&31.9&\textbf{23.0}\\
&& AP 0.5:0.95 (\%)&41.8&39.5&41.8&41.5&41.4\\
            \multirow{2}{*}{Faster R-CNN X101-FPN}
&\multirow{2}{*}{-}&Average Latency ($\mu$s) &37.0&86.9&33.4&28.6&\textbf{20.6}\\
&& AP 0.5:0.95 (\%)&43.0&40.5&43.0&42.7&42.5\\
            \bottomrule
        \end{tabular}
    }
\end{table}

MaxpoolNMS \cite{Cai_2019_CVPR} and ASAP-NMS \cite{tripathi2020asap} are only applicable to the first stage of two-stage detectors, while the problem we are discussing is more general, so we do not include them in our comparison. We compare our methods with PSRR-MaxpoolNMS \cite{Zhang_2021_CVPR}, which is applicable to anchor-based models. We conduct experiments on anchor-based models (Faster R-CNN and YOLOv5) using MS COCO 2017, and the results are shown in Table \ref{tab:psrr-anchor-base}. As we can see, eQSI-NMS achieves the lowest latency while maintaining a favorable trade-off with mAP. However, PSRR-MaxpoolNMS experiences a $1\sim 2\%$ mAP loss in the Faster R-CNN and YOLOv5 models.

% This is likely due to the following reasons:

% \begin{itemize}
%     \item Some hyperparameters need adjustment when using the MS COCO dataset, as MS COCO is more complex compared to PASCAL VOC, with a richer variety of categories and broader scenes. Therefore, the number of scales and ratios may need to be increased to suit the MS COCO dataset. In contrast, our proposed methods do not require additional parameters beyond those used in original NMS, making them more applicable to general cases.
%     \item The PSRR-MaxpoolNMS paper does not specify how the input image size and the parameters $W$ and $H$, which represent the width and height of the image, respectively, are determined. In our implementation, we set these dimensions to $640\times 640$ to ensure compatibility with all images in the MS COCO 2017 dataset.
% \end{itemize}

In the case of Faster R-CNN, the latency performance of PSRR-MaxpoolNMS is not competitive. This is because PSRR-MaxpoolNMS requires 8 max-pooling operations, which, although not affecting the algorithm's complexity, introduces a large constant factor that hampers efficiency when the number of bounding boxes is small (e.g., the average number of bounding boxes in the three Faster R-CNN models is less than 300). However, it performs well when the number of bounding boxes is large (e.g., YOLOv5-S has an average of 2898 bounding boxes). This demonstrates that the speedup of PSRR-MaxpoolNMS is highly dependent on the degree of parallelism, whereas our methods directly reduce computational overhead (see Figure~\ref{fig:apendix:1} in Appendix~\ref{appedix:more-results}), making it hardware-agnostic and suitable for resource-constrained edge devices.

\section{Related Work}

NMS algorithm is widely used in object detection tasks \cite{greedynms_edge_detection, YOLO, FasterRCNN}. Original NMS \cite{greedynms_human_dectection} operates on a greedy principle, suppressing bounding boxes with an Intersection over Union (IOU) higher than a given threshold, starting from the ones with the highest confidence scores. On one hand, numerous improvements have been made to NMS to achieve higher mAP in certain scenarios \cite{softNMS, weightedNMS, varvoting, adaptiveNMS, NOHNMS, LearningNMS, Seq2SeqNMS}. On the other hand, some research focuses on enhancing speed. Fast NMS \cite{yoloact} improves NMS efficiency by avoiding the sequential processing of bounding boxes that need to be suppressed, making it more conducive to parallel computing and thus speeding up the process, though it may slightly reduce accuracy compared to original NMS. Cluster-NMS \cite{ClusterNMS}, employs matrix operations and iterative processing, running the Fast NMS algorithm in each iteration to accelerate the original NMS without compromising accuracy. MaxpoolNMS \cite{Cai_2019_CVPR} and ASAP-NMS \cite{tripathi2020asap} take into account the setting of "anchors" in the region proposal network (RPN) of two-stage detectors. MaxpoolNMS maps anchors of different sizes onto several score maps and performs spatial max-pooling on these score maps to avoid calculating IOUs, thereby improving the speed of NMS. ASAP-NMS eliminates some boxes with relatively small IOUs by precomputing the IOUs between the current box and neighboring anchors. PSRR-MaxpoolNMS \cite{Zhang_2021_CVPR} improves upon MaxpoolNMS by introducing Relationship Recovery, which addresses the issue of score map mismatch that may arise in MaxpoolNMS, enabling PSRR-MaxpoolNMS to be used in the second stage of two-stage detectors.
CUDA NMS by torchvision \cite{TorchVision} is a CUDA implementation of the original NMS, leveraging a GPU to accelerate computation-intensive tasks, though it cannot be used in scenarios without a GPU.

\section{Conclusion}

In this paper, we systematically analyze the NMS algorithm from a graph theory perspective and discover strong connections between NMS, directed graph topological sorting, dynamic programming, and weak connected components. Through these analyses, we first propose QSI-NMS, a fast divide-and-conquer algorithm with negligible loss, and its extended version, eQSI-NMS, achieves the state-of-the-art complexity $\mathcal{O}(n\log n)$. Additionally, starting from the sparsity of graphs, we design BOE-NMS, which considers the locality suppression feature of NMS, optimizes the NMS algorithm at a constant level, and maintains precision. Furthermore, we introduce NMS-Bench, the first end-to-end benchmark integrating bounding box datasets, NMS benchmarking methods, and evaluations, facilitating NMS research for researchers. Finally, we conducted experiments on NMS-Bench, and the experimental results validated our theory, demonstrating the superiority of our algorithms.

\paragraph{Acknowledgements}

This work was supported in part by the National Natural Science Foundation of China under Grant 62192781, Grant 62172326 and Grant 62137002, and in part by the Project of China Knowledge Centre for Engineering Science and Technology.

%%%%%%%%%%%%%%%%%%%%%%%%%%%%%%%%%%%%%%%%%%%%%%%%%%%%%%%%%%%%
\bibliographystyle{plain}
\bibliography{ref.bib}

%%%%%%%%%%%%%%%%%%%%%%%%%%%%%%%%%%%%%%%%%%%%%%%%%%%%%%%%%%%%

\clearpage
\appendix
\renewcommand{\appendixpagename}{Appendix}
\appendixpage
\startcontents[sections]
% \startlist{lof}
% \startlist{lot}
\section*{Table of Contents}
\vspace{-15pt}\noindent\hrulefill

\printcontents[sections]{l}{1}{\setcounter{tocdepth}{2}}

\noindent\hrulefill

\clearpage

\section{Preliminaries}\label{pre}

\subsection{Graph Theory}

\paragraph{Notations and Terminology} A directed graph $\mathcal{G}$, also called a digraph, is an ordered pair $(\mathcal{V},\mathcal{E})$ consisting of a nonempty set $\mathcal{V}$ of nodes, a set $\mathcal{E}$ of arcs. An arc, also called an arrow, is an ordered pair $(v, u)$ of nodes, where $v,u\in\mathcal{V}$. $v$ and $u$ are the head and tail of the arc $(v, u)$, respectively. We say two nodes $v, u$ are adjacent if there exists an arc $e\in\mathcal{E}$ such that $e=(v, u)\vee e=(u, v)$. A direct predecessor of node $v$ is the head of an arc whose tail is $v$, and a direct successor is the tail of an arc whose head is $v$. The in-degree $d^{-}(v)$ of a node $v\in\mathcal{V}$ is the number of arcs with tail $v$ while the out-degree $d^{+}(v)$ is the number of arcs with head $v$. A walk $(v_0, v_1,\ldots,v_k)$ in a directed graph is a sequence of nodes satisfying $(v_i, v_{i+1})\in\mathcal{E}$ for all $i=0,1,\ldots,k-1$. A path in a directed graph is a walk in which all vertices are distinct. A cycle is a path $(v_0,v_1,\ldots,v_k)$ together with the arc $(v_k,v_0)$. We say a graph $\mathcal{G}$ is simple if there are no multi-edges or self-loops in $\mathcal{G}$. $\mathcal{G}=(\mathcal{V},\mathcal{E})$ contains a graph $\mathcal{G}'=(\mathcal{V}', \mathcal{E}')$ if $\mathcal{V}'\subset \mathcal{V}\wedge \mathcal{E}'\subset \mathcal{E}$, we then say $\mathcal{G}'$ is a subgraph of $\mathcal{G}$ and denote $\mathcal{G}'\subset \mathcal{G}$.

We next introduce several key concepts.

\begin{definition}[Directed acyclic graph]
A directed acyclic graph (DAG) is a directed graph without cycles.
\end{definition}

\begin{definition}[Strongly connected]
A directed graph $\mathcal{G} = (\mathcal{V},\mathcal{E})$ is called strongly connected if it contains a path between $v,u$ and a path between $u,v$, for every pair $v, u\in\mathcal{V}$.
\end{definition}
\begin{definition}[Weakly connected]
A directed graph $\mathcal{G}=(\mathcal{V},\mathcal{E})$ is weakly connected if the symmetric graph $\mathcal{G}' = (\mathcal{V},\mathcal{E}')$ with $\mathcal{E}'=\mathcal{E}\cup \{(u, v)\vert (v,u)\in\mathcal{E}\}$ is strongly connected.
\end{definition}
\begin{definition}[Strongly connected component]
For a directed graph, an inclusion-maximal strongly connected subgraph is called strongly connected component (SCC).
\end{definition}
\begin{definition}[Weakly connected component]
For a directed graph, an inclusion-maximal weakly connected subgraph is called weakly connected component (WCC).
\end{definition}

\subsection{Cartesian Tree}

\begin{definition}\label{def:cartesian}
A Cartesian tree for a sequence is a binary tree constructed as follows, 
\begin{itemize}
    \item The root of the tree is the maximum element of the sequence.
    \item Its left and right subtrees are formed by recursively constructing Cartesian tree for the subsequences to the left and right.
\end{itemize}
The basic case is that if the sequence is empty, then the Cartesian tree is also empty.
\end{definition}

During the construction process of a Cartesian tree, the maximum value is selected from the current sequence as the root each time. Thus, the Cartesian tree obeys the max-heap property whose inorder traversal returns the original sequence. The complexity of constructing a Cartesian tree according to Definition~\ref{def:cartesian} is $\mathcal{O}(n^2)$ in the worst-case which is too slow. Fortunately, we have the following proposition:
\begin{proposition}\label{pro:ct_on}
Given a sequence $S$ of length $n$, a Cartesian tree for $S$ can be constructed in $\mathcal{O}(n)$ time.
\end{proposition}
Here, we provide an informal proof. We can establish Proposition~\ref{pro:ct_on} by designing an algorithm with $\mathcal{O}(n)$ complexity. We consider adding elements from $S$ one by one to construct the Cartesian tree. Assume that we have already constructed the Cartesian tree for $S[1:k-1]$. Now, when adding $S[k]$, we observe that $S[k]$ will be the rightmost node of the tree. Furthermore, since Cartesian trees are max-heaps, this implies that the parent of $S[k]$ must be greater than or equal to $S[k]$. Therefore, we need to find the rightmost node $v$ in the Cartesian tree that is greater than or equal to $S[k]$, and assign its right child $u$ as the left child of $S[k]$, updating the right child of $v$ to be $S[k]$. We can implement this algorithm efficiently by maintaining a stack: pop the stack until the stack top is not less than $S[k]$, and then update the tree structure accordingly. After this, push $S[k]$ onto the stack. When $k$ is iterated from $1$ to $n$, the construction of the Cartesian tree is also completed. Since each element is pushed onto and popped from the stack at most once, the overall complexity is $\mathcal{O}(n)$.

\clearpage
\section{Proofs}\label{proofs}

\subsection{Proof of Proposition~\ref{pro:methodology:1}}

\begin{proof}
Assume that there is cycle in the directed graph $\mathcal{G}$ with length $k$, i.e., there exists a sequence
\[
(v_0, v_1, v_2, \ldots, v_{k}), 
\]
where $v_0=v_{k}$ and $\forall i=0,1,\ldots,k-1, (v_i,v_{i+1})\in\mathcal{E}$.

By definition, we know that if $(v,u)\in \mathcal{E}$, then $b_v>b_u$. Therefore,
\[
b_{v_{0}}>b_{v_1}>b_{v_2}>\ldots >b_{v_k}=b_{v_0},
\]
which demonstrates that
\[
b_{v_0}>b_{v_0}.
\]
This is contradictory to the fact that $b_{v_0}$ is a real number. 
\end{proof}

\subsection{Proof of Theorem~\ref{thm:methodology:1}}

\begin{proof}
We first sort the scores in $\mathcal{S}$ in descending order into a sequence $S:(s_{i_1},s_{i_2},\ldots,s_{i_n})$. To prove Theorem~\ref{thm:methodology:1}, we prove that statement $\delta(i_t)=k_{i_t}$ holds for all $t=1,2,\ldots,n$ by mathematical induction.

For $t=1$, $k_{i_1}=1$, that's because no box can suppress $b_{i_1}$ with the maximum score $s_{i_1}$. $\delta(i_1)=1=k_{i_1}$ since the in-degree of $i_1$ is $0$. Therefore, the statement is true for $t=1$.

Assume it is true for integers $1,2,\ldots,t-1,t\geq 2$, then we turn our attention to $t$. If $k_{i_t}=1$, it implies that there does not exist a $p<t$ such that retained $b_{i_p}$ can suppress $b_{i_t}$, i.e., $\text{IOU}(b_{i_t}, b_{i_p})\leq N_t\vee k_{i_p}=0$. Therefore, if there is an arc from $i_p$ to $i_t$, then $\delta(i_p)=k_{i_p}=0$, which indicates $\delta(i_{t})=1$. Similarly, we can prove that $\delta(i_t)=0$ if $k_{i_t}=0$.

By mathematical induction, the statement above is true for $1,2,\ldots,n$.
\end{proof}

\subsection{Proof of Theorem~\ref{thm:methodology:2}}

\begin{proof}
We denote $\vert \mathcal{B}\vert$ by $n$, we prove Theorem~\ref{thm:methodology:2} by mathematical induction.

For $n=1$, the statement is true since there is only one node in both binary trees.

Suppose that the statement holds for all sets with size $1,2,\ldots,t$. Consider any set $\mathcal{B}$ with $n=(t+1)$ elements.

The root of $QT(\mathcal{B})$ corresponds $b_v$ with maximum $s_v$ in $\mathcal{B}$ ($QT(\emptyset)=\emptyset$). And the root of the Cartesian tree corresponds the maximum element of the sequence $B$, which implies it's also $b_v$ indexed by $m$, i.e., $i_m=v$.

According to Definition~\ref{def:methodology:2}, the left child of $QT(\mathcal{B})$'s root is $\mathcal{B}_l$'s root which satisfies
\[
c_u \preceq_{\mathcal{C}} c_v \text{ for all }b_u \in \mathcal{B}_l,
\]
where each $c_u$ is the centroid of the corresponding box $b_u$, and $c_v$ is the centroid of $b_v$.

Notice that $B$ is sorted in ascending order of centroids according to the order $\preceq_{\mathcal{C}}$, which demonstrates that the elements in the set $\mathcal{B}_l$ are the same to that in the sequence $B[1:m-1]$. Formally, $\vert \mathcal{B}_l\vert=m-1$ and $\forall b_v\in\mathcal{B}_l$, there exists one and only one $x\in\{1,2,\ldots,m-1\}$ such that $v=i_{x}$, i.e., $b_v=b_{i_x}$. 

According to the hypothesis, $QT(\mathcal{B}_l)$ is a Cartesian tree for $B[1:m-1]$. Similarly, we can prove $QT(\mathcal{B}_r)$ is a Cartesian tree for $B[m+1:t+1]$. 

Hence, the statement is true for $n=(t+1)$, which completes the proof.
\end{proof}

\subsection{Proof of Theorem~\ref{thm:methodology:3}} 

\begin{proof}
We can take $b$ as the frame of reference. As illustrated in Figure~\ref{fig:apendix:1}, the blue box is $b$, and the green one is $b^*$ which is respectively positioned to the left (Figure~\ref{subfig:1}) or right (Figure~\ref{subfig:2}) of the vertical dashed line, or above (Figure~\ref{subfig:3}) or below (Figure~\ref{subfig:4}) the horizontal dashed line.
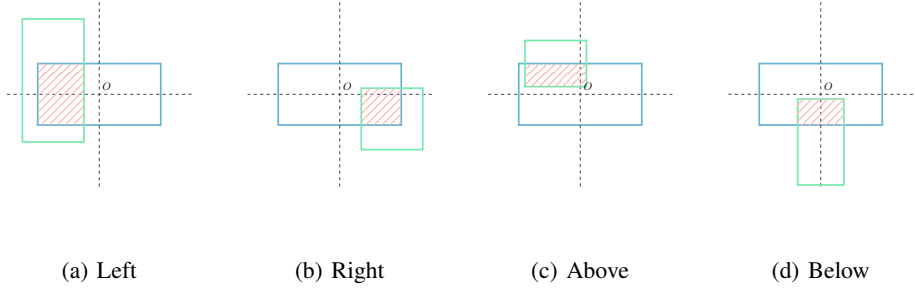
\begin{figure}[ht]
    \centering
    \subfigure[Left]{\label{subfig:1}
        \begin{minipage}{.21\linewidth}
            % \documentclass[tikz, convert, convert={outext=.pdf,
% 		command=\unexpanded{}}]{standalone}
% \usetikzlibrary{arrows.meta,decorations.pathreplacing, patterns, intersections, calc}
% \usepackage{fontspec}
% \setmainfont{Times New Roman}

\resizebox{\linewidth}{!}{
\begin{tikzpicture}
    \definecolor{bbox_c}{HTML}{67B3D2}
    \definecolor{mbox_c}{HTML}{76EAAE}
    \definecolor{hvline_c}{HTML}{354A53}

    \tikzset{
        bbox/.style={
            draw=bbox_c, minimum width=4cm, minimum height=2cm, 
            ultra thick
        },
        mbox/.style={
            draw=mbox_c, ultra thick
        },
        hvline/.style={
            thick, dashed, draw=hvline_c
        },
    }

    \draw [draw=none] (-8.5, -8.5) grid (-1.5, -1.5);

    % x,y axes
    % \draw [dashed, thick] (-10, 0) -- (10, 0);
    % \draw [dashed, thick] (0, -10) -- (0, 10);

    % bbox
    \foreach \v in {-1} {
        \foreach \u in {-1} {
            \node (b{\v}{\u}) [bbox, name path=pathb{\v}{\u}] at (\v * 5, \u * 5) {};
        }
    }

    % h, v lines
    \foreach \v in {-1} {
        \foreach \u in {-1} {
            \draw [hvline] (\v * 8, \u * 5) -- (\v * 2, \u * 5);
            \draw [hvline] (\v * 5, \u * 2) -- (\v * 5, \u * 8);
        }
    }

    % mbox
    % \coordinate (m1o) at (-5.8, 6);
    % \node (m1) [mbox, minimum width=2cm, minimum height=1.5cm, name path=pathm1] at (m1o) {};

    % \path [name intersections={of=pathm1 and pathb{-1}{1}}]
    %     coordinate [] (m1a) at (intersection-1)
    %     coordinate [] (m1b) at (intersection-2);

    % \draw [pattern=north east lines, pattern color=red!30, draw=none] ($(m1o) + (-1, -0.75)$) -- ($(m1o) + (1, -0.75)$) -- (m1b) -- (m1a) -- ($(m1o) + (-1, -0.75)$);

    % \coordinate (m2o) at (5, 3.45);
    % \node (m2) [mbox, minimum width=1.5cm, minimum height=2.8cm, name path=pathm2] at (m2o) {};

    % \path [name intersections={of=pathm2 and pathb{1}{1}}]
    %     coordinate [] (m2a) at (intersection-1)
    %     coordinate [] (m2b) at (intersection-2);

    % \draw [pattern=north east lines, pattern color=red!30, draw=none] ($(m2o) + (-.75, 1.4)$) -- ($(m2o) + (.75, 1.4)$) -- (m2b) -- (m2a) -- ($(m2o) + (-.75, 1.4)$);

    \coordinate (m3o) at (-6.5, -4.55);
    \node (m3) [mbox, minimum width=2cm, minimum height=4cm, name path=pathm3] at (m3o) {};

    \path [name intersections={of=pathm3 and pathb{-1}{-1}}]
        coordinate [] (m3a) at (intersection-1)
        coordinate [] (m3b) at (intersection-2);

    \draw [pattern=north east lines, pattern color=red!30, draw=none] (-7, -6) -- (m3b) -- (m3a) -- (-7, -4);

    % \coordinate (m4o) at (6.7, -5.8);
    % \node (m4) [mbox, minimum width=2cm, minimum height=2cm, name path=pathm4] at (m4o) {};

    % \path [name intersections={of=pathm4 and pathb{1}{-1}}]
    %     coordinate [] (m4a) at (intersection-1)
    %     coordinate [] (m4b) at (intersection-2);

    % \draw [pattern=north east lines, pattern color=red!30, draw=none] (7, -6) -- (m4b) -- ($(m4o) + (-1, 1)$) -- (m4a);

    \foreach \v in {-1} {
        \foreach \u in {-1} {
            \node [] at (\v * 5 + 0.25, \u * 5 + 0.25) {$O$};
        }
    }

\end{tikzpicture}
}
        \end{minipage}
    }
    \subfigure[Right]{\label{subfig:2}
        \begin{minipage}{.21\linewidth}
            % \documentclass[tikz, convert, convert={outext=.pdf,
% 		command=\unexpanded{}}]{standalone}
% \usetikzlibrary{arrows.meta,decorations.pathreplacing, patterns, intersections, calc}
% \usepackage{fontspec}
% \setmainfont{Times New Roman}

\resizebox{\linewidth}{!}{
\begin{tikzpicture}
    \definecolor{bbox_c}{HTML}{67B3D2}
    \definecolor{mbox_c}{HTML}{76EAAE}
    \definecolor{hvline_c}{HTML}{354A53}

    \tikzset{
        bbox/.style={
            draw=bbox_c, minimum width=4cm, minimum height=2cm, 
            ultra thick
        },
        mbox/.style={
            draw=mbox_c, ultra thick
        },
        hvline/.style={
            thick, dashed, draw=hvline_c
        },
    }

    \draw [draw=none] (1.5, -8.5) grid (8.5, -1.5);

    % x,y axes
    % \draw [dashed, thick] (-10, 0) -- (10, 0);
    % \draw [dashed, thick] (0, -10) -- (0, 10);

    % bbox
    \foreach \v in {1} {
        \foreach \u in {-1} {
            \node (b{\v}{\u}) [bbox, name path=pathb{\v}{\u}] at (\v * 5, \u * 5) {};
        }
    }

    % h, v lines
    \foreach \v in {1} {
        \foreach \u in {-1} {
            \draw [hvline] (\v * 8, \u * 5) -- (\v * 2, \u * 5);
            \draw [hvline] (\v * 5, \u * 2) -- (\v * 5, \u * 8);
        }
    }

    % mbox
    % \coordinate (m1o) at (-5.8, 6);
    % \node (m1) [mbox, minimum width=2cm, minimum height=1.5cm, name path=pathm1] at (m1o) {};

    % \path [name intersections={of=pathm1 and pathb{-1}{1}}]
    %     coordinate [] (m1a) at (intersection-1)
    %     coordinate [] (m1b) at (intersection-2);

    % \draw [pattern=north east lines, pattern color=red!30, draw=none] ($(m1o) + (-1, -0.75)$) -- ($(m1o) + (1, -0.75)$) -- (m1b) -- (m1a) -- ($(m1o) + (-1, -0.75)$);

    % \coordinate (m2o) at (5, 3.45);
    % \node (m2) [mbox, minimum width=1.5cm, minimum height=2.8cm, name path=pathm2] at (m2o) {};

    % \path [name intersections={of=pathm2 and pathb{1}{1}}]
    %     coordinate [] (m2a) at (intersection-1)
    %     coordinate [] (m2b) at (intersection-2);

    % \draw [pattern=north east lines, pattern color=red!30, draw=none] ($(m2o) + (-.75, 1.4)$) -- ($(m2o) + (.75, 1.4)$) -- (m2b) -- (m2a) -- ($(m2o) + (-.75, 1.4)$);

    % \coordinate (m3o) at (-6.5, -4.55);
    % \node (m3) [mbox, minimum width=2cm, minimum height=4cm, name path=pathm3] at (m3o) {};

    % \path [name intersections={of=pathm3 and pathb{-1}{-1}}]
    %     coordinate [] (m3a) at (intersection-1)
    %     coordinate [] (m3b) at (intersection-2);

    % \draw [pattern=north east lines, pattern color=red!30, draw=none] (-7, -6) -- (m3b) -- (m3a) -- (-7, -4);

    \coordinate (m4o) at (6.7, -5.8);
    \node (m4) [mbox, minimum width=2cm, minimum height=2cm, name path=pathm4] at (m4o) {};

    \path [name intersections={of=pathm4 and pathb{1}{-1}}]
        coordinate [] (m4a) at (intersection-1)
        coordinate [] (m4b) at (intersection-2);

    \draw [pattern=north east lines, pattern color=red!30, draw=none] (7, -6) -- (m4b) -- ($(m4o) + (-1, 1)$) -- (m4a);

    \foreach \v in {1} {
        \foreach \u in {-1} {
            \node [] at (\v * 5 + 0.25, \u * 5 + 0.25) {$O$};
        }
    }

\end{tikzpicture}
}
        \end{minipage}
    }
    \subfigure[Above]{\label{subfig:3}
        \begin{minipage}{.21\linewidth}
            % \documentclass[tikz, convert, convert={outext=.pdf,
% 		command=\unexpanded{}}]{standalone}
% \usetikzlibrary{arrows.meta,decorations.pathreplacing, patterns, intersections, calc}
% \usepackage{fontspec}
% \setmainfont{Times New Roman}

\resizebox{\linewidth}{!}{
\begin{tikzpicture}
    \definecolor{bbox_c}{HTML}{67B3D2}
    \definecolor{mbox_c}{HTML}{76EAAE}
    \definecolor{hvline_c}{HTML}{354A53}

    \tikzset{
        bbox/.style={
            draw=bbox_c, minimum width=4cm, minimum height=2cm, 
            ultra thick
        },
        mbox/.style={
            draw=mbox_c, ultra thick
        },
        hvline/.style={
            thick, dashed, draw=hvline_c
        },
    }

    \draw [draw=none] (-8.5, 1.5) grid (-1.5, 8.5);

    % x,y axes
    % \draw [dashed, thick] (-10, 0) -- (10, 0);
    % \draw [dashed, thick] (0, -10) -- (0, 10);

    % bbox
    \foreach \v in {-1} {
        \foreach \u in {1} {
            \node (b{\v}{\u}) [bbox, name path=pathb{\v}{\u}] at (\v * 5, \u * 5) {};
        }
    }

    % h, v lines
    \foreach \v in {-1} {
        \foreach \u in {1} {
            \draw [hvline] (\v * 8, \u * 5) -- (\v * 2, \u * 5);
            \draw [hvline] (\v * 5, \u * 2) -- (\v * 5, \u * 8);
        }
    }

    % mbox
    \coordinate (m1o) at (-5.8, 6);
    \node (m1) [mbox, minimum width=2cm, minimum height=1.5cm, name path=pathm1] at (m1o) {};

    \path [name intersections={of=pathm1 and pathb{-1}{1}}]
        coordinate [] (m1a) at (intersection-1)
        coordinate [] (m1b) at (intersection-2);

    \draw [pattern=north east lines, pattern color=red!30, draw=none] ($(m1o) + (-1, -0.75)$) -- ($(m1o) + (1, -0.75)$) -- (m1b) -- (m1a) -- ($(m1o) + (-1, -0.75)$);

    % \coordinate (m2o) at (5, 3.45);
    % \node (m2) [mbox, minimum width=1.5cm, minimum height=2.8cm, name path=pathm2] at (m2o) {};

    % \path [name intersections={of=pathm2 and pathb{1}{1}}]
    %     coordinate [] (m2a) at (intersection-1)
    %     coordinate [] (m2b) at (intersection-2);

    % \draw [pattern=north east lines, pattern color=red!30, draw=none] ($(m2o) + (-.75, 1.4)$) -- ($(m2o) + (.75, 1.4)$) -- (m2b) -- (m2a) -- ($(m2o) + (-.75, 1.4)$);

    % \coordinate (m3o) at (-6.5, -4.55);
    % \node (m3) [mbox, minimum width=2cm, minimum height=4cm, name path=pathm3] at (m3o) {};

    % \path [name intersections={of=pathm3 and pathb{-1}{-1}}]
    %     coordinate [] (m3a) at (intersection-1)
    %     coordinate [] (m3b) at (intersection-2);

    % \draw [pattern=north east lines, pattern color=red!30, draw=none] (-7, -6) -- (m3b) -- (m3a) -- (-7, -4);

    % \coordinate (m4o) at (6.7, -5.8);
    % \node (m4) [mbox, minimum width=2cm, minimum height=2cm, name path=pathm4] at (m4o) {};

    % \path [name intersections={of=pathm4 and pathb{1}{-1}}]
    %     coordinate [] (m4a) at (intersection-1)
    %     coordinate [] (m4b) at (intersection-2);

    % \draw [pattern=north east lines, pattern color=red!30, draw=none] (7, -6) -- (m4b) -- ($(m4o) + (-1, 1)$) -- (m4a);

    \foreach \v in {-1} {
        \foreach \u in {1} {
            \node [] at (\v * 5 + 0.25, \u * 5 + 0.25) {$O$};
        }
    }

\end{tikzpicture}
}
        \end{minipage}
    }
    \subfigure[Below]{\label{subfig:4}
        \begin{minipage}{.21\linewidth}
            % \documentclass[tikz, convert, convert={outext=.pdf,
% 		command=\unexpanded{}}]{standalone}
% \usetikzlibrary{arrows.meta,decorations.pathreplacing, patterns, intersections, calc}
% \usepackage{fontspec}
% \setmainfont{Times New Roman}

\resizebox{\linewidth}{!}{
\begin{tikzpicture}
    \definecolor{bbox_c}{HTML}{67B3D2}
    \definecolor{mbox_c}{HTML}{76EAAE}
    \definecolor{hvline_c}{HTML}{354A53}

    \tikzset{
        bbox/.style={
            draw=bbox_c, minimum width=4cm, minimum height=2cm, 
            ultra thick
        },
        mbox/.style={
            draw=mbox_c, ultra thick
        },
        hvline/.style={
            thick, dashed, draw=hvline_c
        },
    }

    \draw [draw=none] (1.5, 1.5) grid (8.5, 8.5);

    % x,y axes
    % \draw [dashed, thick] (-10, 0) -- (10, 0);
    % \draw [dashed, thick] (0, -10) -- (0, 10);

    % bbox
    \foreach \v in {1} {
        \foreach \u in {1} {
            \node (b{\v}{\u}) [bbox, name path=pathb{\v}{\u}] at (\v * 5, \u * 5) {};
        }
    }

    % h, v lines
    \foreach \v in {1} {
        \foreach \u in {1} {
            \draw [hvline] (\v * 8, \u * 5) -- (\v * 2, \u * 5);
            \draw [hvline] (\v * 5, \u * 2) -- (\v * 5, \u * 8);
        }
    }

    % mbox
    % \coordinate (m1o) at (-5.8, 6);
    % \node (m1) [mbox, minimum width=2cm, minimum height=1.5cm, name path=pathm1] at (m1o) {};

    % \path [name intersections={of=pathm1 and pathb{-1}{1}}]
    %     coordinate [] (m1a) at (intersection-1)
    %     coordinate [] (m1b) at (intersection-2);

    % \draw [pattern=north east lines, pattern color=red!30, draw=none] ($(m1o) + (-1, -0.75)$) -- ($(m1o) + (1, -0.75)$) -- (m1b) -- (m1a) -- ($(m1o) + (-1, -0.75)$);

    \coordinate (m2o) at (5, 3.45);
    \node (m2) [mbox, minimum width=1.5cm, minimum height=2.8cm, name path=pathm2] at (m2o) {};

    \path [name intersections={of=pathm2 and pathb{1}{1}}]
        coordinate [] (m2a) at (intersection-1)
        coordinate [] (m2b) at (intersection-2);

    \draw [pattern=north east lines, pattern color=red!30, draw=none] ($(m2o) + (-.75, 1.4)$) -- ($(m2o) + (.75, 1.4)$) -- (m2b) -- (m2a) -- ($(m2o) + (-.75, 1.4)$);

    % \coordinate (m3o) at (-6.5, -4.55);
    % \node (m3) [mbox, minimum width=2cm, minimum height=4cm, name path=pathm3] at (m3o) {};

    % \path [name intersections={of=pathm3 and pathb{-1}{-1}}]
    %     coordinate [] (m3a) at (intersection-1)
    %     coordinate [] (m3b) at (intersection-2);

    % \draw [pattern=north east lines, pattern color=red!30, draw=none] (-7, -6) -- (m3b) -- (m3a) -- (-7, -4);

    % \coordinate (m4o) at (6.7, -5.8);
    % \node (m4) [mbox, minimum width=2cm, minimum height=2cm, name path=pathm4] at (m4o) {};

    % \path [name intersections={of=pathm4 and pathb{1}{-1}}]
    %     coordinate [] (m4a) at (intersection-1)
    %     coordinate [] (m4b) at (intersection-2);

    % \draw [pattern=north east lines, pattern color=red!30, draw=none] (7, -6) -- (m4b) -- ($(m4o) + (-1, 1)$) -- (m4a);

    \foreach \v in {1} {
        \foreach \u in {1} {
            \node [] at (\v * 5 + 0.25, \u * 5 + 0.25) {$O$};
        }
    }

\end{tikzpicture}
}
        \end{minipage}
    }
    \caption{Four positions of $b^*$ relative to $b$.}
    \label{fig:apendix:1}
\end{figure}
The intersections are filled with red north east lines. Hence, we have
\begin{align*}
    \text{IOU}(b^*,b)&=\text{IOU}(b,b^*)\\
    &=\frac{\text{Area}(\text{red})}{\text{Union}(b,b^*)}\\
    &\leq \frac{1/2\text{Area}(b)}{\text{Area}(b)}\\
    &=\frac{1}{2}.
\end{align*}
\end{proof}

\subsection{Proof of Theorem~\ref{thm:methodology:4}}

We prove Theorem 4 by demonstrating the inequalities in the following two lemmas.

\begin{lemma}\label{lemma:ineq1}
Given positive real numbers $\theta$ and $\beta$, for any $x_1, x_2, y_1, y_2\in\mathbb{R}$, the following inequality holds:
\[ 
\theta (x_2 - x_1) + \beta (y_2 - y_1) \geq (\theta + \beta) (\min\{x_2,y_2\}-\max\{x_1,y_1\}).
\]
\end{lemma}
\begin{lemma}\label{leamm:ineq2}
Given a positive real number $\gamma$, for any $x_1,x_2,y_1,y_2\in\mathbb{R}$ such that
\[
(\gamma + 1)x_2 + (1-\gamma)x_1\leq y_1+y_2, 
\]
the following inequality holds:
\[
x_2+y_2-x_1-y_1 \geq (2+\gamma)(\min\{x_2,y_2\}-\max\{x_1,y_1\}).
\]
\end{lemma}

\begin{proof}[Proof of Lemma~\ref{lemma:ineq1}]

Notice that $\min\{x_2, y_2\} \leq x_2$ and $\min\{x_2, y_2\} \leq y_2$, we have:
\[
\min\{x_2, y_2\} \leq \frac{\theta}{\theta + \beta} x_2 + \frac{\beta}{\theta + \beta} y_2.
\]
For $\max\{x_1,y_1\}$, we have:
\[
\max\{x_1,y_1\} \geq \frac{\theta}{\theta + \beta} x_1 + \frac{\beta}{\theta + \beta} y_1.
\]
Therefore, 
\begin{align*}
    RHS&=(\theta + \beta) (\min\{x_2,y_2\}-\max\{x_1,y_1\})\\
    &\leq (\theta + \beta) (\frac{\theta}{\theta + \beta} x_2 + \frac{\beta}{\theta + \beta} y_2 - \frac{\theta}{\theta + \beta} x_1 - \frac{\beta}{\theta + \beta} y_1)\\
    &=(\theta + \beta) (\frac{\theta}{\theta + \beta}(x_2-x_1) + \frac{\beta}{\theta + \beta}(y_2-y_1))\\
    &=\theta (x_2 - x_1) + \beta (y_2 - y_1)\\
    &=LHS.
\end{align*}
\end{proof}

\begin{proof}[Proof of Lemma~\ref{leamm:ineq2}]
Similar to the approach used in the proof above, for $\max\{x_1,y_1\}$, we have
\[
\max\{x_1,y_1\}\geq \frac{\gamma}{2+\gamma}x_1+\frac{2}{2+\gamma}y_1.
\]
Notice that $\min\{x_2,y_2\}\leq x_2$, then we have
\begin{align*}
    LHS&=x_2-x_1+(y_1+y_2)-2y_1\\
    &\geq x_2-x_1 + (\gamma + 1)x_2+(1-\gamma)x_1 -2y_1\\
    &=(2+\gamma)x_2-\gamma x_1-2y_1\\
    &=(2+\gamma)(x_2-\frac{\gamma}{2+\gamma}x_1-\frac{2}{2+\gamma}y_1)\\
    &\geq (2+\gamma)(\min\{x_2,y_2\}-\max\{x_1,y_1\})\\
    &=RHS.
\end{align*}
\end{proof}

We next use these two inequalities to prove Theorem~\ref{thm:methodology:4}.

\begin{proof}[Proof of Theorem~\ref{thm:methodology:4}]
For convenience, we denote a bounding box $b$ by a 4-tuple $(x^{(b)}_{lt}, x^{(b)}_{rb}, y^{(b)}_{lt}, y^{(b)}_{rb})$, where $(x^{(b)}_{lt},y^{(b)}_{lt})$ and $(x^{(b)}_{rb},y^{(b)}_{rb})$ represent the left-top and right-bottom corners of $b$, respectively. This says
\[
\begin{cases}
    x^{(b)}_{lt}\leq x^{(b)}_{rb},\\
    y^{(b)}_{lt}\leq y^{(b)}_{rb}.
\end{cases}
\]
Given bounding boxes $b^*$ and $b$, their intersection can be calculated as follows:
\begin{align*}
\text{Inter}(b^*, b)&=\max\{0, \min\{x^{(b^*)}_{rb}, x^{(b)}_{rb}\} - \max\{x^{(b^*)}_{lt}, x^{(b)}_{lt}\}\} \\
&\times \max\{0, \min\{y^{(b^*)}_{rb}, y^{(b)}_{rb}\} - \max\{y^{(b^*)}_{lt}, y^{(b)}_{lt}\}\}.
\end{align*}
Here, $b^*$ is represented by $(x_{lt}^{(b^*)}, x_{rb}^{(b^*)}, y_{lt}^{(b^*)}, y_{rb}^{(b^*)})$ and $b$ by $(x_{lt}^{(b)}, x_{rb}^{(b)}, y_{lt}^{(b)}, y_{rb}^{(b)})$. Let
\[
\begin{cases}
I_x=\min\{x^{(b^*)}_{rb}, x^{(b)}_{rb}\} - \max\{x^{(b^*)}_{lt}, x^{(b)}_{lt}\},\\
I_y=\min\{y^{(b^*)}_{rb}, y^{(b)}_{rb}\} - \max\{y^{(b^*)}_{lt}, y^{(b)}_{lt}\}.
\end{cases}
\]
We observe that if $I_x \leq 0$ or $I_y \leq 0$, then $\text{Inter}(b^*, b) = 0$, resulting in $\text{IOU} = 0$. This is a trivial case, and Theorem~\ref{thm:methodology:4} holds. Therefore, we only need to consider the case where $I_x > 0$ and $I_y > 0$. In this case, 
\[
\text{Inter}(b^*, b) = I_x I_y.
\]
Then the union can be expressed as:
\begin{align*}
\text{Union}(b^*, b) &= \text{Area}(b^*)+\text{Area}(b) - \text{Inter}(b^*,b)\\
&=L_{x}^{(b^*)}L_y^{(b^*)} + L_x^{(b)}L_y^{(b)} - I_x I_y, 
\end{align*}
where
\[
\begin{cases}
    L_x^{(b^*)}=x^{(b^*)}_{rb}-x^{(b^*)}_{lt};\\
    L_y^{(b^*)}=y^{(b^*)}_{rb}-y^{(b^*)}_{lt};\\
    L_x^{(b)}=x^{(b)}_{rb}-x^{(b)}_{lt};\\
    L_y^{(b)}=y^{(b)}_{rb}-y^{(b)}_{lt},
\end{cases}
\]
represent the width and height of $b^*$ and $b$, respectively. We then have the following inequality holds:
\begin{equation}\label{inq:thm4:1}
\frac{L_x^{(b^*)}L_y^{(b^*)}+L_x^{(b)}L_y^{(b)}}{I_xI_y}\geq \frac{1}{N_t}+1.
\end{equation}
Let $\theta=L_x^{(b^*)}>0,\beta=L_x^{(b)}>0$, according to Lemma~\ref{lemma:ineq1} and $I_y>0$, then we have
\begin{equation}\label{inq:thm4:2}
\frac{L_x^{(b^*)}L_y^{(b^*)}+L_x^{(b)}L_y^{(b)}}{I_y}\geq L_x^{(b^*)}+L_x^{(b)}.
\end{equation}
Furthermore, according to the proof of Theorem~\ref{thm:methodology:3}, $\alpha(b^*,\nicefrac{1}{N_t}-1)$ is respectively positioned to the left or right of the vertical dashed line, or above or below the horizontal dashed line (see Figure~\ref{fig:apendix:1}). Without loss of generality, we can assume that $\alpha(b^*,\nicefrac{1}{N_t}-1)$ is to the left of vertical dashed line passing through the centroid of $b$. This means:
\begin{align*}
    \frac{(x_{lt}^{(b)}+x_{rb}^{(b)})}{2}&\geq x^{(b^*)}_{c} + s\times \vert x^{(b^*)}_{rb}-x^{(b^*)}_{c}\vert\\
    &=(x_{lt}^{(b^*)}+x_{rb}^{(b^*)})/2 + (1/N_t-1)\times (x_{rb}^{(b^*)}-x_{lt}^{(b^*)})/2\\
    &=\frac{(1/N_t)x_{rb}^{(b^*)}+(2-1/N_t)x_{lt}^{(b^*)}}{2}.
\end{align*}
let $\gamma$ be $\nicefrac{1}{N_t}-1>0$, according to Lemma~\ref{leamm:ineq2} and $I_x>0$, then we have
\begin{equation}\label{inq:thm4:3}
\frac{L_{x}^{(b^*)}+L_x^{(b)}}{I_x}\geq 2+(\frac{1}{N_t}-1)=\frac{1}{N_t}+1.
\end{equation}
Combining inequalities (\ref{inq:thm4:2}) and (\ref{inq:thm4:3}), inequality (\ref{inq:thm4:1}) is proven. 

Finally, according to the calculation method of IOU, we have:
\begin{align*}
    \text{IOU}(b^*,b)&=\frac{\text{Inter}(b^*,b)}{\text{Union}(b^*,b)}\\
    &=\frac{I_xI_y}{L_x^{(b^*)}L_y^{(b^*)}+L_x^{(b)}L_y^{(b)} - I_xI_y}\\
    &=\frac{1}{\frac{L_x^{(b^*)}L_y^{(b^*)}+L_x^{(b)}L_y^{(b)}}{I_xI_y}-1}\\
    &\leq \frac{1}{(\frac{1}{N_t}+1)-1}\\
    &=N_t.
\end{align*}
\end{proof}

\subsection{Proof of Proposition \ref{pro:appendix:2}}\label{prf:pro:appendix:2}

\begin{proof}
Since a point can be considered as an interval of length $0$, we can construct the set $\mathcal{M}'$ in $\mathcal{O}(n)$ time:
$$
\mathcal{M}' = \{[m_i, m_i] \mid m_i \in \mathcal{M}\}.
$$
Suppose there is an algorithm $A$ that can solve problem $X$. This means:
$$
\mathcal{Q}_{X} = A(\mathcal{I}, [l, r]).
$$
Then, we can spend $\mathcal{O}(1)$ time to call $A$ once, using $\mathcal{M}'$ and $[l, r]$ as its inputs. Therefore:
$$
\mathcal{Q}_{Y} = A(\mathcal{M}', [l, r]).
$$
Therefore, problem $Y$ can be reduced to $X$ in polynomial time, i.e., $Y\leq_{P}X$.
\end{proof}

\clearpage
\section{Discussion}\label{discussion}

From a graph theory perspective, we revisited the NMS algorithm and discovered statistical properties of the weakly connected components in graph $\mathcal{G}$: they are numerous overall but small in size locally. Leveraging these two characteristics, we proposed QSI-NMS and BOE-NMS. We then validated the efficiency and accuracy of QSI-NMS and BOE-NMS on NMS-Bench, showing significant improvements over the original NMS, parallelized Fast NMS, and Cluster-NMS. However, there are several areas for improvement in our work.

First, our NMS algorithms can be combined with other accuracy-enhancing NMS methods, such as Soft-NMS \cite{softNMS}, to address the negligible accuracy loss introduced by QSI-NMS and eQSI-NMS. Since our algorithms serve as general frameworks, allowing other methods to be implemented by modifying the dynamic programming equation. Second, our algorithms can be further parallelized to improve efficiency. QSI-NMS uses a divide-and-conquer recursive strategy, enabling parallel processing of different subproblems, while BOE-NMS can adopt the approach of Fast NMS, where each box computes local IOUs in parallel. Third, the distribution characteristics of bounding boxes can be further studied to obtain more detailed information about graph $\mathcal{G}$, aiding in the analysis of accuracy loss in Fast NMS, QSI-NMS, and eQSI-NMS. These can be explored in future work.

In the latter part of this section, we delve deeper into our work. In Subsection~\ref{discussion:1}, we conduct a thorough graph theory analysis of Fast NMS and Cluster-NMS; in Subsection~\ref{discussion:psrr}, we compare our methods with MaxpoolNMS, ASAP-NMS, and PSRR-MaxpoolNMS; in Subsection~\ref{discussion:qsi-case-study}, we explain and analyze the reasons behind the slight mAP drop of QSI-NMS; in Subsection~\ref{discussion:qsi-eqsi}, we explore some implementation details of QSI-NMS; and in Subsection~\ref{discussion:boe-nms}, we discuss the advantages of the BOE-NMS implementation. 

\subsection{Discussion of Fast NMS and Cluster-NMS}\label{discussion:1}

\paragraph{Discussion of Fast NMS} In Fast NMS \cite{yoloact}, the computation of IOU is parallelized, and when the IOU exceeds a threshold $N_t$, the bounding box with the higher confidence score always suppresses the one with the lower confidence score. This allows the results for all bounding boxes to be computed in parallel without depending on the results of previous boxes. Formally, in Fast NMS, the set $\mathcal{B}$ is first sorted into $B$ by confidence scores in descending order, and then an $n\times n$ matrix $\mathbf{Y}$ is defined as follows:
\[
y_{i,j}=
\begin{cases}
\text{IOU}(b_i,b_j) & \text{if } i < j;\\
0 &\text{otherwise}.
\end{cases}
\]
For a bounding box $b_j$, if there exists a bounding box $b_i$ with higher confidence ($i < j$) such that $y_{i,j} > N_t$, then $b_j$ is not retained; otherwise, it is retained. Since $N_t$ effectively classifies all IOUs into two categories, for ease of analysis, we define a matrix $\mathbf{X}$ where:
\[
x_{i,j}=
\begin{cases}
1 & \text{if } i < j \text{ and } y_{i,j} > N_t;\\
0 &\text{otherwise}.
\end{cases}
\]
Thus, we have
\[
k_j=\lnot \left(\bigvee_{i<j} x_{i,j}\right).
\]
We can observe that $\mathbf{X}$ is the adjacency matrix of $\mathcal{G}$. Fast NMS can also obtain the same result using dynamic programming, but the dynamic programming equation is as follows:
\[
\delta(u) =
\begin{cases}
0 & \text{if } d^{-}(u) > 0;\\
1 & \text{otherwise}.
\end{cases}
\]
It is evident that Fast NMS suppresses more bounding boxes compared to original NMS, which explains the loss in mAP observed with Fast NMS. Additionally, this demonstrates that Fast NMS can achieve its results without the need for topological sorting, explaining why Fast NMS is parallelizable. However, the actual implementation of Fast NMS has a time complexity of $\Theta(n^2)$, which is, in fact, greater than that of original NMS. In original NMS, suppressed boxes are not used in subsequent IOU calculations, whereas in Fast NMS, every box calculates IOUs with all preceding boxes. This highlights the heavy reliance of Fast NMS on efficient parallel computing.

\paragraph{Discussion of Cluster-NMS} In Cluster-NMS \cite{ClusterNMS}, the issue of accuracy loss is addressed by iterating Fast NMS multiple times. We use a binary vector $\boldsymbol{r}$ to represent the result of applying Fast NMS to $\mathbf{X}$, denoted as $\boldsymbol{r}=F(\mathbf{X})$, where:
$$
\boldsymbol{r}_j=\lnot \left(\bigvee_{i<j} x_{i,j}\right).
$$
The iterative process of Cluster-NMS is as follows:
\begin{enumerate}
\item $\mathbf{X}^{(0)}=\mathbf{X}$;
\item $\boldsymbol{r}^{(t)}=F(\mathbf{X}^{(t)})$, for $t \geq 0$;
\item $\mathbf{X}^{(t+1)}=\mathrm{diag}(\boldsymbol{r}^{(t)})\times\mathbf{X}$;
\item Iterate until $\boldsymbol{r}^{(t)}$ converges to $\boldsymbol{r}^*$, at which point $\Vert \boldsymbol{r}^{(t)} - \boldsymbol{r}^* \Vert = 0$.
\end{enumerate}

In step 3, the iterative method for $\mathbf{X}^{(t+1)}$ involves left-multiplying $\mathbf{X}$ by a diagonal $01$ matrix, meaning rows corresponding to $1$ are retained and those corresponding to $0$ are discarded. Formally expressed as follows:
\[
x_{i,j}^{(t+1)}=
\begin{cases}
x_{i,j} & \text{if } \boldsymbol{r}_i^{(t)} = 1; \\
0 & \text{otherwise}.
\end{cases}
\]
This shows that only boxes not suppressed in the current iteration can cause a suppression effect in the next iteration. Thus, bounding boxes incorrectly suppressed in this round will be re-evaluated with truly retained boxes in the next iteration to determine if they should indeed be suppressed. 

\cite{ClusterNMS} proves that its results are equivalent to original NMS and usually requires only a few iterations. To discuss the number of iterations, \cite{ClusterNMS} defines a cluster:

\begin{definition}[cluster in Cluster-NMS \cite{ClusterNMS}]
A subset $\mathcal{U} = \{b_{j_1}, b_{j_2}, \ldots, b_{j_{\vert \mathcal{U} \vert}}\}$ of $\mathcal{B}$ is a cluster if and only if for all $b_{j_t} \in \mathcal{U}$, there exists $i \in \{j_1, j_2, \ldots, j_{\vert\mathcal{U}\vert}\} \setminus \{j_t\}$ such that \(\text{IOU}(b_{j_t}, b_{i}) > N_t\) and for all $b \in \mathcal{B} \setminus \mathcal{U}$, $\text{IOU}(b_{j_t}, b) \leq N_t$.
\end{definition}

\cite{ClusterNMS} proves that the number of iterations does not exceed the size of the largest cluster. 

Analyzing Cluster-NMS from the perspective of graph $\mathcal{G}$, each iteration essentially determines whether all current nodes with an in-degree of $0$ should be retained, i.e., the value of $\delta(\cdot)$; then it traverses all outgoing arcs of some node $v$ with in-degree 0 to decide if a successor node $u$ should be suppressed, updating $\delta(u)$ as follows:
\[
\delta(u) \leftarrow \delta(u) \wedge \lnot \delta(v).
\]
Finally, all outgoing arcs are deleted, and the next iteration begins. Hence, the essence of Cluster-NMS is parallel topological sorting within each WCC. In fact, we can see that the boxes in a cluster correspond to a WCC in $\mathcal{G}$. This explains why the number of iterations until convergence does not exceed the size of the largest cluster: in each iteration, at least one node with an in-degree of $0$ is expanded and added to the topological sort within each WCC, reducing its size by at least one per iteration. Although Cluster-NMS is correct, the matrix $\mathbf{X}$ obtained by parallel IOU computation already encodes all the information of $\mathcal{G}$, which indicates a single dynamic programming can quickly produce results identical to original NMS.

\subsection{Discussion of MaxpoolNMS, ASAP-NMS, and PSRR-MaxpoolNMS}\label{discussion:psrr}

\paragraph{Discussion of MaxpoolNMS and ASAP-NMS}

MaxpoolNMS \cite{Cai_2019_CVPR} and ASAP-NMS \cite{tripathi2020asap} take into account the information of pre-defined ``anchors'' in the RPN and leverage the locality suppression characteristics of NMS to achieve impressive performance improvements. However, they have the following limitations:

\begin{itemize}
    \item MaxpoolNMS and ASAP-NMS can only be used in the first stage of two-stage detectors. Our methods, however, can be used in any stage of any detector because we address the most general case (see Section~\ref{sec:pro-def}).
    \item The complexity of MaxpoolNMS is $\mathcal{O}(n_s n_r\lfloor \frac{W}{\beta} \rceil \lfloor \frac{H}{\beta} \rceil + n \log n)$, where $n_s$ and $n_r$ represent the number of anchor box scales and ratios, respectively. The complexity of ASAP-NMS is $\mathcal{O}(n^2)$. Neither of these methods are more efficient than eQSI-NMS.
    \item These methods involve many manually defined hyperparameters and are complex to implement, which limits their generalization across different models and datasets. In contrast, our methods require no additional parameters beyond those in original NMS and are easy to implement.
    \item These methods are not rigorous and can lead to a certain degree of mAP degradation, whereas BOE-NMS is rigorously proven to cause no mAP loss.
\end{itemize}

\paragraph{Discussion of PSRR-MaxpoolNMS} 

\begin{figure}[ht]
    \centering
    \subfigure[]{\label{subfig:psrr-maxpoolnms:a}
        \begin{minipage}{.48\linewidth}
            \centering
        \includegraphics[width=\linewidth]{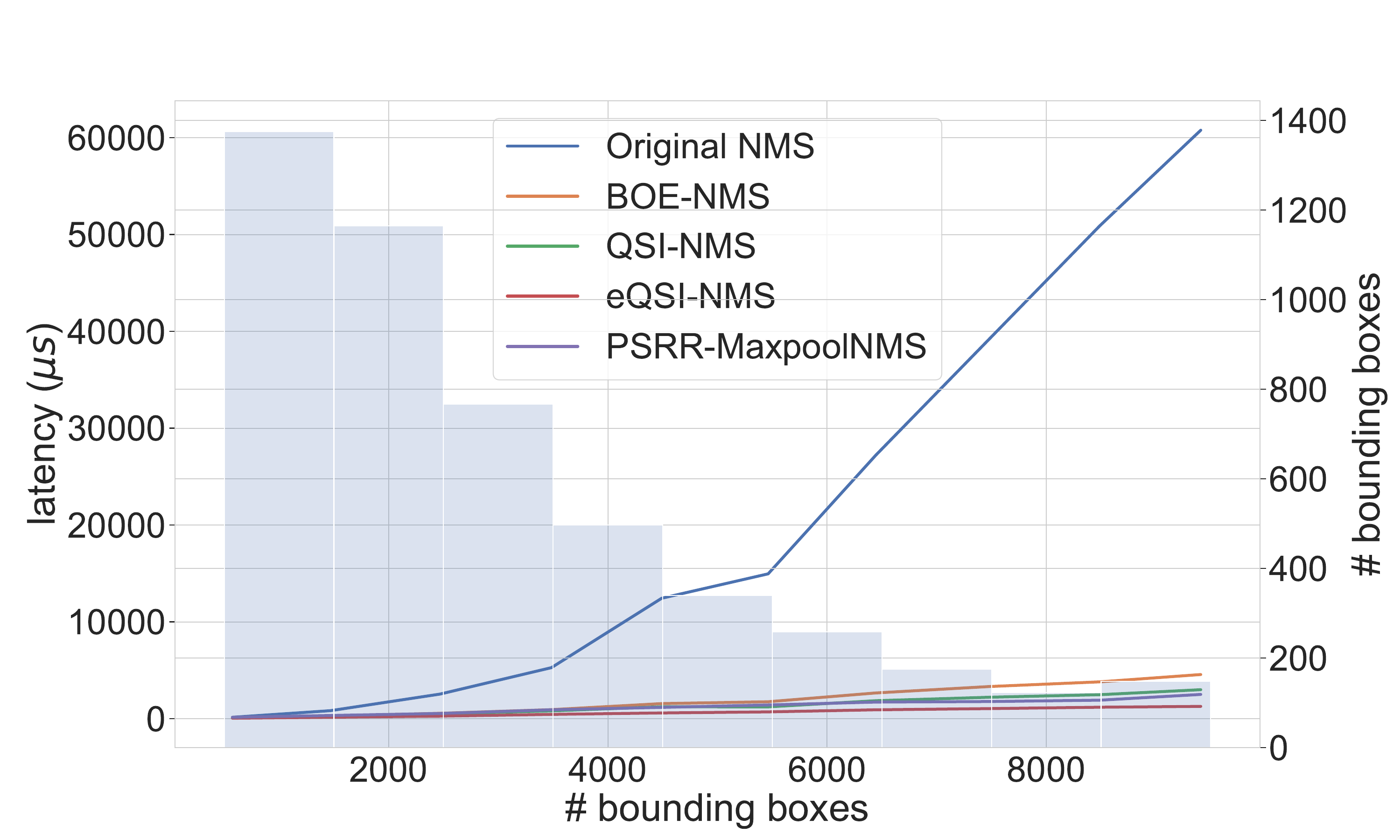}
        \end{minipage}
    }
    \subfigure[]{\label{subfig:psrr-maxpoolnms:b}
        \begin{minipage}{.48\linewidth}
            \centering
            \includegraphics[width=\linewidth]{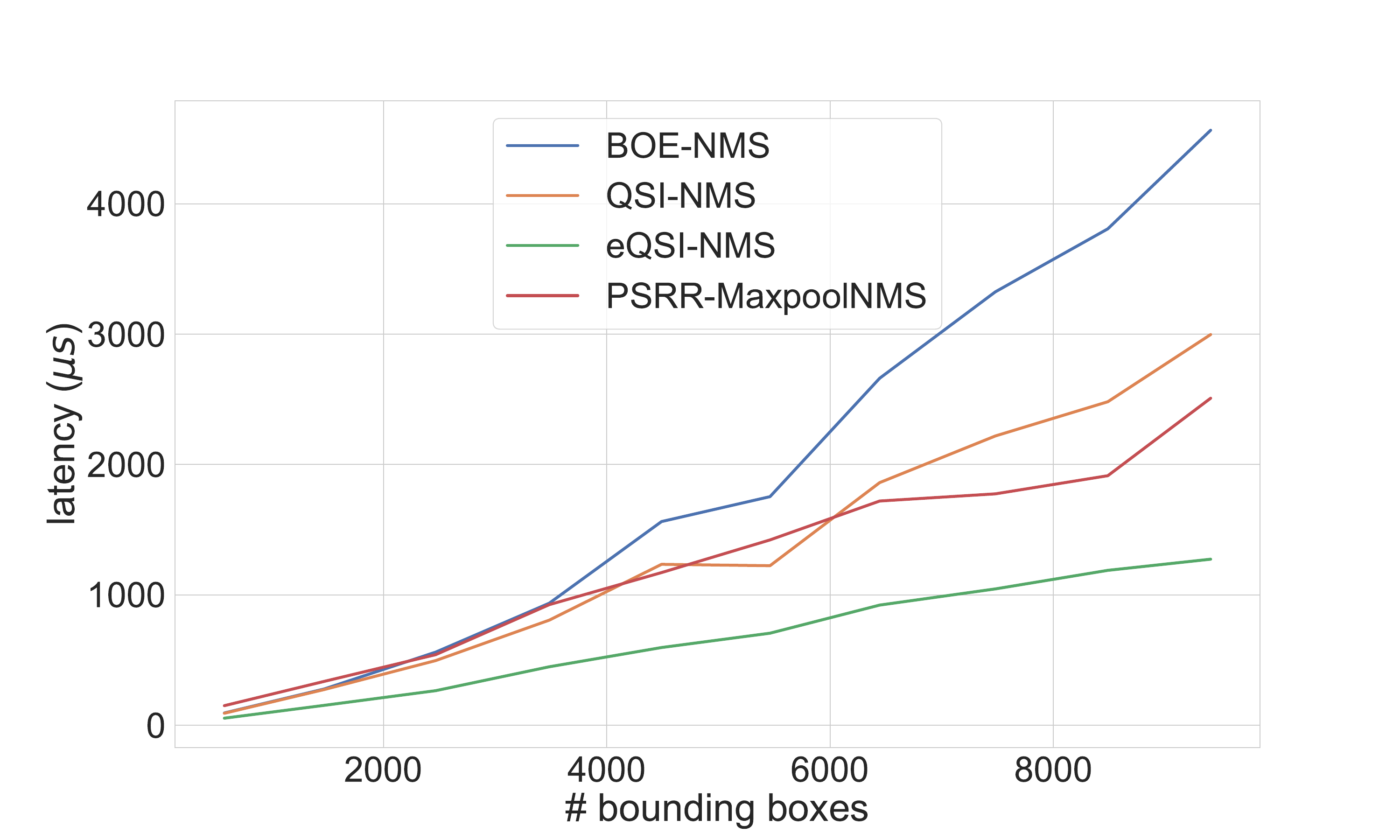}
        \end{minipage}
    }
    \caption{The line plot of the runtime of different methods as the number of bounding boxes varies in YOLOv5-N. \ref{subfig:psrr-maxpoolnms:a} The histogram with a bin width of 1000 representing the number of bounding boxes in each interval. The input images are divided into 10 intervals based on the number of bounding boxes: $(0,1000], (1000,2000], \ldots, (9000,10000]$. The line plot is drawn with the average number of boxes per interval as the x-coordinate and the average time cost of the NMS algorithms as the y-coordinate. \ref{subfig:psrr-maxpoolnms:b} The line plot of the runtime of different methods without original NMS.}
    \label{fig:psrr-maxpoolnms}
\end{figure}

PSRR-MaxpoolNMS \cite{Zhang_2021_CVPR} introduces Relationship Recovery to address the issue of regression box and score map mismatch in MaxpoolNMS, allowing it to be used at any stage of all anchor-based detectors, and claims to achieve a time complexity of $\mathcal{O}(n)$. However, we do not believe that PSRR-MaxpoolNMS outperforms our methods for the following two reasons:

First, we implement PSRR-MaxpoolNMS and conduct efficiency comparisons. To better illustrate the impact of bounding box count on runtime, we perform experiments on YOLOv5-N, which has the highest number of bounding boxes, as shown in Figure~\ref{subfig:psrr-maxpoolnms:a}. Original NMS has the highest time cost due to its quadratic growth. For a clearer comparison between our methods and PSRR-MaxpoolNMS, we exclude original NMS, as shown in Figure~\ref{subfig:psrr-maxpoolnms:b}. As the number of boxes increases, PSRR-MaxpoolNMS is faster than BOE-NMS and QSI-NMS but consistently slower than eQSI-NMS.

Second, strictly speaking, the complexity of PSRR-MaxpoolNMS is not $\mathcal{O}(n)$. PSRR-MaxpoolNMS requires prior knowledge of the input image size and generates confidence score maps related to the size of the image. This aspect is not considered in the complexity analysis (whereas our methods are designed and analyzed independently of the image size). During the Channel Recovery stage of PSRR-MaxpoolNMS, the complexity of computing the nearest distances for channel mapping is $\mathcal{O}(n \times n_s \times n_r)$, where $n_s$ and $n_r$ represent the number of anchor box scales and ratios, respectively. These quantities vary with different datasets and detectors and increase as image size and object count increase. The remaining stages: Spatial Recovery, Pyramid MaxpoolNMS, and Shifted MaxpoolNMS can all be completed in $\mathcal{O}(n)$ time. Thus, the overall complexity of PSRR-MaxpoolNMS is $\mathcal{O}(n_s n_r\lfloor \frac{W}{\beta} \rceil \lfloor \frac{H}{\beta} \rceil + n_s n_r n)$.

\subsection{Discussion on mAP loss of QSI-NMS}\label{discussion:qsi-case-study}

In QSI-NMS, we use a divide-and-conquer strategy, which means that bounding boxes in different subproblems do not affect each other. In some special cases, QSI-NMS may assign nodes from the same WCC to different subproblems, potentially causing some nodes that should have been suppressed to be retained.

\begin{figure}[ht]
    \centering
    \subfigure[]{\label{subfig:qsi-case-study:1a}
        \begin{minipage}{.48\linewidth}
        \includegraphics[width=\linewidth]{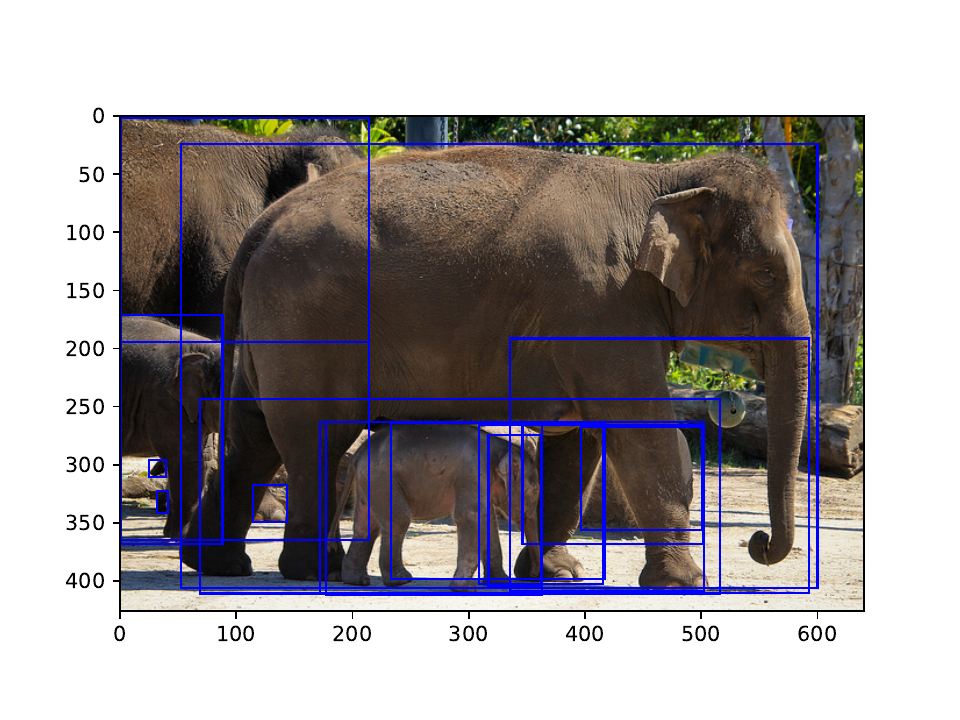}
        \end{minipage}
    }
    \subfigure[]{\label{subfig:qsi-case-study:1b}
        \begin{minipage}{.48\linewidth}
        \includegraphics[width=\linewidth]{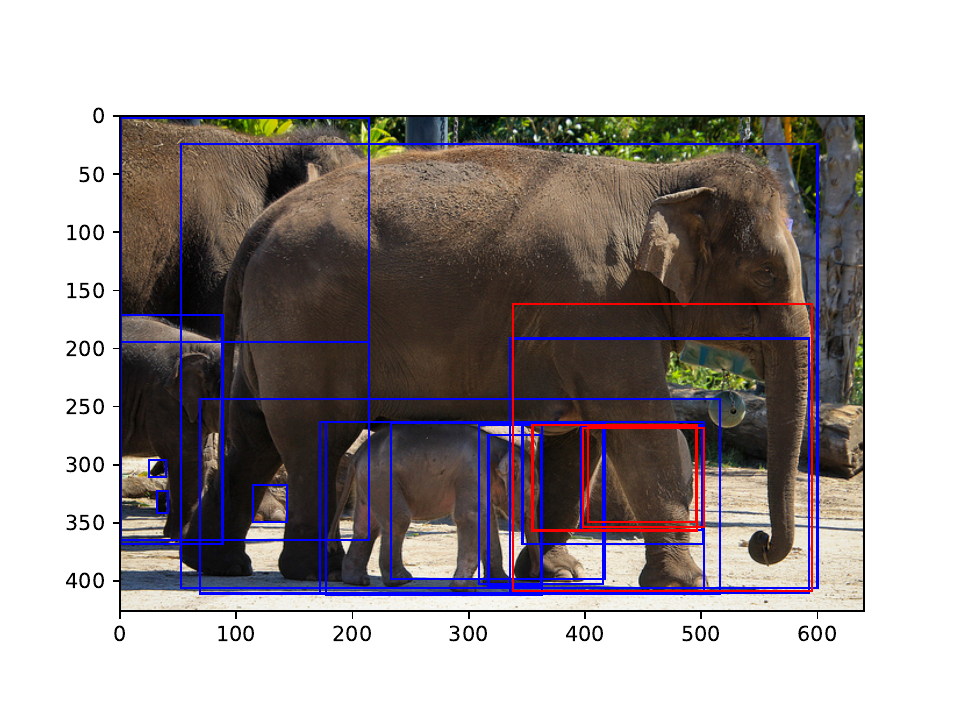}
        \end{minipage}
    }
    \caption{The output bounding boxes of original NMS (\ref{subfig:qsi-case-study:1a}) and QSI-NMS (\ref{subfig:qsi-case-study:1b}) in YOLOv8-M on the MS COCO 2017 image ``000000057027.jpg''.}
    \label{fig:qsi-case-study:1}
\end{figure}

We provide a case study with results from the YOLOv8-M model on MS COCO 2017. In Figure~\ref{subfig:qsi-case-study:1a}, the blue boxes represent the outputs of original NMS/BOE-NMS, while Figure~\ref{subfig:qsi-case-study:1b} shows the outputs of QSI-NMS, with red boxes indicating additional boxes retained by QSI-NMS. It can be seen that QSI-NMS retains four additional boxes.

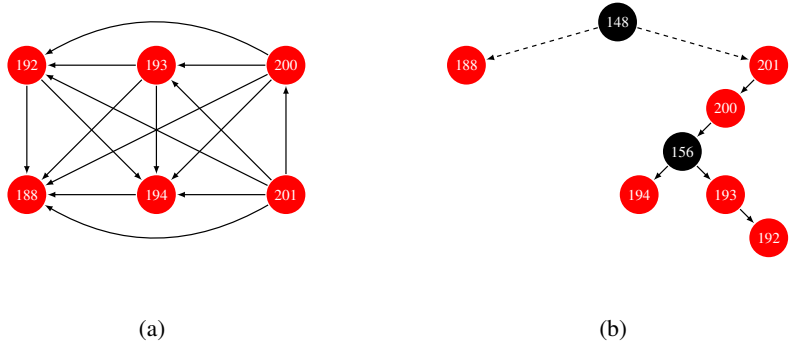
\begin{figure}[ht]
    \centering
    \subfigure[]{\label{subfig:qsi-case-study:2a}
        \begin{minipage}{.42\linewidth}
            \centering
        \resizebox{.8\linewidth}{!}{
    \begin{tikzpicture}
        \definecolor{node1_c}{HTML}{000000}
        \definecolor{node0_c}{HTML}{FFFFFF}
        \definecolor{node2_c}{HTML}{FF0000}
    
        \tikzset{
            point/.style = {
                minimum size=.5cm, circle, ultra thick
            },
            arrow/.style = {
                thick, -latex
            },
            dash-arrow/.style = {
                thick, -latex, dashed
            }
        }
    
        \draw [draw=none] (-4, 4.5) grid (4, 10.5);
    
        % \node (148) [point, fill=node1_c, text=white] at (0, 10) {148};
        \node (192) [point, fill=node2_c, text=white] at (-3, 9) {192};
        \node (188) [point, fill=node2_c, text=white] at (-3, 6) {188};
        
        \node (193) [point, fill=node2_c, text=white] at (0, 9) {193};
        \node (194) [point, fill=node2_c, text=white] at (0, 6) {194};
        
        \node (200) [point, fill=node2_c, text=white] at (3, 9) {200};
        \node (201) [point, fill=node2_c, text=white] at (3, 6) {201};
        % \node (156) [point, fill=node1_c, text=white] at (2, 7) {156};
    
        \draw [arrow] (192) -- (188);
        \draw [arrow] (192) -- (194);
    
        \draw [arrow] (193) -- (188);
        \draw [arrow] (193) -- (192);
        \draw [arrow] (193) -- (194);
    
        \draw [arrow] (194) -- (188);
    
        \draw [arrow] (200) -- (188);
        \draw [arrow] (200) [out=150, in=30] to (192);
        \draw [arrow] (200) -- (193);
        \draw [arrow] (200) -- (194);
    
        \draw [arrow] (201) [out=-150, in=-30] to (188);
        \draw [arrow] (201) -- (192);
        \draw [arrow] (201) -- (193);
        \draw [arrow] (201) -- (194);
        \draw [arrow] (201) -- (200);
    
    \end{tikzpicture}
}
        \end{minipage}
    }
    \subfigure[]{\label{subfig:qsi-case-study:2b}
        \begin{minipage}{.42\linewidth}
            \centering
        \resizebox{.8\linewidth}{!}{
    \begin{tikzpicture}
        \definecolor{node1_c}{HTML}{000000}
        \definecolor{node0_c}{HTML}{FFFFFF}
        \definecolor{node2_c}{HTML}{FF0000}
    
        \tikzset{
            point/.style = {
                minimum size=.5cm, circle, ultra thick
            },
            arrow/.style = {
                thick, -latex
            },
            dash-arrow/.style = {
                thick, -latex, dashed
            }
        }
    
        \draw [draw=none] (-4, 4.5) grid (4, 10.5);
    
        \node (148) [point, fill=node1_c, text=white] at (0, 10) {148};
        \node (188) [point, fill=node2_c, text=white] at (-3.5, 9) {188};
        \node (201) [point, fill=node2_c, text=white] at (3.5, 9) {201};
        \node (200) [point, fill=node2_c, text=white] at (2.5, 8) {200};
        \node (156) [point, fill=node1_c, text=white] at (1.5, 7) {156};
        \node (194) [point, fill=node2_c, text=white] at (0.5, 6) {194};
        \node (193) [point, fill=node2_c, text=white] at (2.5, 6) {193};
        \node (192) [point, fill=node2_c, text=white] at (3.5, 5) {192};
        
        \draw [dash-arrow] (148) -- (201);
        \draw [dash-arrow] (148) -- (188);
        \draw [arrow] (201) -- (200);
        \draw [arrow] (200) -- (156);
        \draw [arrow] (156) -- (194);
        \draw [arrow] (156) -- (193);
        \draw [arrow] (193) -- (192);
    
    \end{tikzpicture}
}
        \end{minipage}
    }
    \caption{WCC in graph $\mathcal{G}$ contains node $188$ (\ref{subfig:qsi-case-study:2a}), along with a partial structure of the QSI-tree (\ref{subfig:qsi-case-study:2b}).}
    \label{fig:qsi-case-study:2}
\end{figure}

For example, consider the box/node numbered 188. The WCC containing this node is shown in Figure~\ref{subfig:qsi-case-study:2a}. All other nodes in the WCC would suppress node 188, but when we use $\preceq_{M}$ to define the partitioning criterion, node 188 ends up in a different subproblem than other nodes, as shown in Figure~\ref{subfig:qsi-case-study:2b}. The figure shows a partial structure of the QSI-tree: solid lines indicate parent-child relationships, and dashed lines indicate ancestor-descendant relationships. The red nodes are nodes from the WCC, while the black node 148 is the lowest common ancestor (LCA) of nodes 188 and 201; node 156 is the LCA of nodes 194 and 193. Since each node in this WCC can only be suppressed by its red ancestor nodes, node 188 is not suppressed. However, node 194 is still suppressed because node 201 is its ancestor.

This example highlights the core of QSI-NMS design: the pivot selection and the partitioning criterion. If we choose these two appropriately, the accuracy loss of QSI-NMS can be negligible. In our algorithm design: pivot selection chooses the most representative nodes (with the highest confidence scores), so node 194 is correctly suppressed by node 201 even after being placed in a different subproblem from node 193. The partitioning criterion aims to assign nodes from the same WCC to the same subproblem as much as possible, which helps reduce cases like node 188 being incorrectly retained. We also discuss other partitioning criteria in Appendix~\ref{discussion:qsi-eqsi}.

\subsection{Further Discussion of QSI-NMS and eQSI-NMS}\label{discussion:qsi-eqsi}
    Algorithm~\ref{alg:qsi-nms} and Algorithm~\ref{alg:eqsi-nms} describe QSI-NMS and eQSI-NMS respectively, which we use C++ to implement in NMS-Bench. 

    In the QSI-NMS and eQSI-NMS, we define the order $\preceq_{\mathcal{C}}$, where $\mathcal{C}$ represents the set of centroids of the original boxes. The preorder $\preceq_{\mathcal{C}}$ is defined on $\mathbb{R}^2$. Different preorders result in slight variations in the final graph $\tilde{\mathcal{G}}$, which in turn cause minor differences in the NMS results. Therefore, we conduct comparative experiments on different orders. 

    We explore three classical orders ($\preceq_{L}, \preceq_{M}, \preceq_{E}$), defined as follows:
    \begin{align*}
            (x_1, y_1) \preceq_{L} (x_2, y_2) &\Leftrightarrow (x_1 < x_2) \lor (x_1 = x_2 \land y_1 \leq y_2)\\
    (x_1, y_1) \preceq_{M} (x_2, y_2) &\Leftrightarrow \lvert x_1 \rvert + \lvert y_1 \rvert \leq \lvert x_2 \rvert + \lvert y_2 \rvert\\
    (x_1, y_1) \preceq_{E} (x_2, y_2) &\Leftrightarrow \sqrt{x_1^2 + y_1^2} \leq \sqrt{x_2^2 + y_2^2}\\
    \end{align*}
    We conduct tests on the MS COCO 2017 using different weights of YOLOv8, with the results shown in Table~\ref{tab:order}. We find orders $\preceq_{M}$ and $\preceq_{E}$ outperform $\preceq_{L}$. 
    
    \begin{table}[htbp]
    \centering
    \caption{AP 0.5:0.95 ($\%$) of QSI-NMS and eQSI-NMS under Different Orders on MS COCO 2017}
    \label{tab:order}
        \begin{tabular}{@{}cccccc@{}}
            \toprule
            \textbf{Model}           & \textbf{Size}      & \textbf{Methods} & \textbf{$\preceq_{L}$} & \textbf{$\preceq_{M}$} & $\preceq_{E}$ \\
            \midrule
            \multirow{10}{*}{YOLOv8} & \multirow{2}{*}{N} & QSI-NMS          & 37.0                               & 37.1                               & 37.1                      \\
                                     &                    & eQSI-NMS         & 36.8                               & 36.9                               & 36.9                      \\
                                     & \multirow{2}{*}{S} & QSI-NMS          & 44.5                               & 44.6                               & 44.6                      \\
                                     &                    & eQSI-NMS         & 44.4                               & 44.5                               & 44.5                      \\
                                     & \multirow{2}{*}{M} & QSI-NMS          & 49.9                               & 50.0                               & 50.0                      \\
                                     &                    & eQSI-NMS         & 49.7                               & 49.9                               & 49.9                      \\
                                     & \multirow{2}{*}{L} & QSI-NMS          & 52.5                               & 52.7                               & 52.7                      \\
                                     &                    & eQSI-NMS         & 52.3                               & 52.5                               & 52.5                      \\
                                     & \multirow{2}{*}{X} & QSI-NMS          & 53.6                               & 53.8                               & 53.8                      \\
                                     &                    & eQSI-NMS         & 53.4                               & 53.6                               & 53.6                      \\
            \bottomrule
        \end{tabular}
\end{table}
    
    In the case where weakly connected components are independent of each other, $\preceq_{M}$ and $\preceq_{E}$ can better maintain the neighborhood consistency between the boxes. In contrast, $\preceq_{L}$ may order the boxes of different weakly connected components closer together, thereby disrupting the hierarchical nature of QSI-NMS and resulting in more accuracy loss. Therefore, when selecting the order, it is necessary to consider whether the construction of the order can retain the positional information of the original boxes as much as possible.

  % In our experiments, we use $\preceq_{\mathcal{C}}^2$ as the order of $\mathcal{C}$.

\subsection{Further Discussion of BOE-NMS}\label{discussion:boe-nms}
An intuitive conclusion is that if two bounding boxes do not overlap at all, their IOU is necessarily 0. This is a weaker conclusion than Theorems 3 and 4, but it is actually more challenging to implement.

Without loss of generality, we consider the one-dimensional case. Given a set of $n$ intervals $\mathcal{I}=\{[l_1,r_1],[l_2,r_2],\ldots,[l_n,r_n]\}$, where $l_i \leq r_i$ for $i=1,2,\ldots,n$.

We define the following two problems:

\begin{definition}[Problem $X$]
Given $\mathcal{I}$ and an arbitrary interval $[l,r]$ where $l \leq r$, find all intervals in $\mathcal{I}$ that intersect with $[l,r]$, i.e., determine the set $\mathcal{Q}_{X}$:
\[
\mathcal{Q}_{X} = \{i \mid 1 \leq i \leq n \wedge [l,r] \cap [l_i, r_i] \neq \emptyset\}.
\]
\end{definition}

\begin{definition}[Problem $Y$]
Given $\mathcal{I}$ and an arbitrary interval $[l, r]$ where $l \leq r$, find all intervals in $\mathcal{I}$ whose midpoints lie within $[l, r]$, i.e., determine the set $\mathcal{Q}_{Y}$:
\[ 
\mathcal{Q}_{Y}=\{i \mid 1 \leq i \leq n \wedge l \leq \frac{l_i+r_i}{2} \leq r\}.
\]
\end{definition}

In problem $X$, we need to find all intervals in $\mathcal{I}$ that intersect with the current interval, while in problem $Y$, we need to find all intervals whose midpoints lie within the current interval. In Problem $Y$, we only need to consider the set of midpoints, which can be represented as:
\[ 
\mathcal{M}=\{m_i \mid m_i=\frac{l_i+r_i}{2}, \forall i=1,2,\ldots,n\}.
\]

For problem $X$, if $i \in \mathcal{Q}_{X}$, meaning that interval $[l_i,r_i]$ intersects with $[l,r]$, then we have
\[ 
l_i \leq r \wedge r_i \geq l.
\]

For problem $Y$, if $i \in \mathcal{Q}_{Y}$, then we have
\[ 
l \leq m_i \leq r.
\]

Therefore, we find that problem $X$ is more complex than problem $Y$ because, in problem $X$, we need to maintain the partial order of both endpoints, whereas in Problem $Y$, we only need to maintain the partial order of the midpoints. Formally, we have the following proposition:

\begin{proposition}\label{pro:appendix:2}
    $Y\leq_{P} X.$
\end{proposition}

We prove Proposition~\ref{pro:appendix:2} in Appendix~\ref{prf:pro:appendix:2}. Proposition~\ref{pro:appendix:2} indicates that problem $Y$ can be reduced to problem $X$. Therefore, $X$ is at least as hard as problem $Y$. Through the above analysis, we can see that although this intuitive conclusion seems straightforward, it is a more difficult problem than determining whether a point lies within a bounding box. Actually, problem $X$ can essentially be equivalent to a 2D plane point filtering problem. Some algorithms can solve this problem by maintaining data structures with large constants, such as persistent segment trees. However, these methods are complex and offer limited optimization. For BOE-NMS, the comparison between points and segments is quite special. We can use sorting and preprocessing of the point set, and then use binary search based on monotonicity to determine the point set corresponding to the query interval. If the size of the point set corresponding to the query interval is $k$, it requires $\mathcal{O}(k + \log(n))$ time complexity to obtain the result, avoiding redundant traversal of invalid points. In the other word, our proposed method BOE-NMS more profoundly exploits the properties of weakly connected components.

\clearpage
\section{Experimental Details}\label{exp-details}
\subsection{More Information about NMS-Bench}
Table~\ref{tab:COCO_number_of_box} and Table~\ref{tab:OI_number_of_box} respectively present the number of original bounding boxes after inferences of different models on the MS COCO 2017 and Open Images V7 datasets. A larger number of bounding boxes indicates weaker filtering capabilities of the model, leading to longer post-processing times required for NMS.
\begin{table}[htbp]
    \centering
    \caption{Number of Bounding Boxes on MS COCO 2017}
    \label{tab:COCO_number_of_box}
    
    \begin{tabular}{@{}ccc@{}}
        \toprule
        \textbf{Model}          & \textbf{Size} & \textbf{Number of Bounding Boxes} \\
        \midrule
        \multirow{5}{*}{YOLOv8} & N             & 4,040,118                      \\
                                & S             & 3,052,794                      \\
                                & M             & 2,627,367                      \\
                                & L             & 2,058,360                      \\
                                & X             & 1,955,594                      \\
        \midrule
        \multirow{5}{*}{YOLOv5} & N             & 14,489,236                     \\
                                & S             & 9,871,207                      \\
                                & M             & 9,051,451                      \\
                                & L             & 7,986,223                      \\
                                & X             & 7,227,635                      \\
        \midrule
        Faster R-CNN R50-FPN    &    -           & 1,256,090                      \\
        Faster R-CNN R101-FPN   &     -          & 1,181,563                      \\
        Faster R-CNN X101-FPN   &      -         & 1,068,974                      \\
        \bottomrule
    \end{tabular}
    \end{table}
    
\begin{table}[htbp]
    \centering
    \caption{Number of Bounding Boxes on Open Images V7}
    \label{tab:OI_number_of_box}
        \begin{tabular}{@{}ccc@{}}
            \toprule
            \textbf{Model}          & \textbf{Size} & \textbf{Number of Bounding Boxes} \\
            \midrule
            \multirow{5}{*}{YOLOv8} & N             & 50,389,365                     \\
                                    & S             & 43,062,828                     \\
                                    & M             & 39,894,713                     \\
                                    & L             & 38,126,739                     \\
                                    & X             & 37,113,747                     \\
            \bottomrule
        \end{tabular}
\end{table}
\subsection{Experimental Environment and Settings}
Our experimental environment is shown as the Table ~\ref{tab:expenv}.

\begin{table}[htbp]
\caption{Experimental Environment}
\label{tab:expenv}
\centering
\begin{tabular}{@{}ll@{}}
\toprule
\textbf{Component} & \textbf{Specification} \\
            \midrule

\multicolumn{2}{@{}l@{}}{\textbf{CPU}} \\
            \midrule

Model & Intel Xeon Gold 6226 \\
Total Cores & 12 \\
Total Threads & 24 \\
Max Turbo Frequency & 3.70 GHz \\
            \midrule

\multicolumn{2}{@{}l@{}}{\textbf{GPU}} \\
            \midrule

Model & NVIDIA RTX 4090 $\times 1$\\
VRAM & 24 GB GDDR6X \\
\bottomrule
\end{tabular}
\end{table}

For the hyperparameter settings, we set the NMS threshold $N_t$ to 0.7 in our experiments.
\subsection{More Results}\label{appedix:more-results}
\paragraph{Experiments in Torchvision Library} To compare with the CUDA NMS from torchvision \cite{TorchVision}, we implement our methods as C++ operators under the torchvision library. We then fairly replace the different NMS operator modules for testing. We test on the MS COCO 2017 using different weights of YOLOv8. The experimental setup is the same as previously described, and we set bench size as 20. The experimental results are shown in the Table~\ref{tab:tc}. This demonstrates that our methods provide performance improvements even when compared to highly optimized parallel implementations.

\begin{table}[htbp] 
    \centering
    \caption{NMS Methods Performance under Torchvision Implementation}
    \label{tab:tc}
    \begin{tabular}{@{}ccccccc@{}}
        \toprule
        \textbf{Model}           & \textbf{Size}      & \textbf{Target} & \textbf{CUDA NMS} & \textbf{BOE-NMS} & \textbf{QSI-NMS} & \textbf{eQSI-NMS} \\
        \midrule
        \multirow{10}{*}{YOLOv8} & \multirow{2}{*}{N} & Average Latency ($\mu$s)       & 343.4             & 204.2            & 185.2            & \textbf{136.3}    \\
                                 &                    & AP 0.5:0.95 (\%)   & 37.4              & 37.4             & 37.3             & 37.1              \\
                                 & \multirow{2}{*}{S} & Average Latency ($\mu$s)       & 302.1             & 158.3            & 160.1            & \textbf{109.4}    \\
                                 &                    & AP 0.5:0.95 (\%)   & 45.0              & 45.0             & 44.8             & 44.7              \\
                                 & \multirow{2}{*}{M} & Average Latency ($\mu$s)       & 301.0             & 136.7            & 141.4            & \textbf{101.3}    \\
                                 &                    & AP 0.5:0.95 (\%)   & 50.3              & 50.3             & 50.2             & 50.1              \\
                                 & \multirow{2}{*}{L} & Average Latency ($\mu$s)       & 284.4             & 109.1            & 111.7            & \textbf{85.2}     \\
                                 &                    & AP 0.5:0.95 (\%)   & 53.0              & 53.0             & 52.8             & 52.7              \\
                                 & \multirow{2}{*}{X} & Average Latency ($\mu$s)       & 284.1             & 109.9            & 106.5            & \textbf{83.5}     \\
                                 &                    & AP 0.5:0.95 (\%)   & 54.0              & 54.0             & 53.9             & 53.8              \\
        \bottomrule
    \end{tabular}
\end{table}

\begin{figure}[htbp]
    \centering
    \subfigure[]{
        \label{sub2:a}
        \begin{minipage}{.48\linewidth}
        \includegraphics[width=\linewidth]{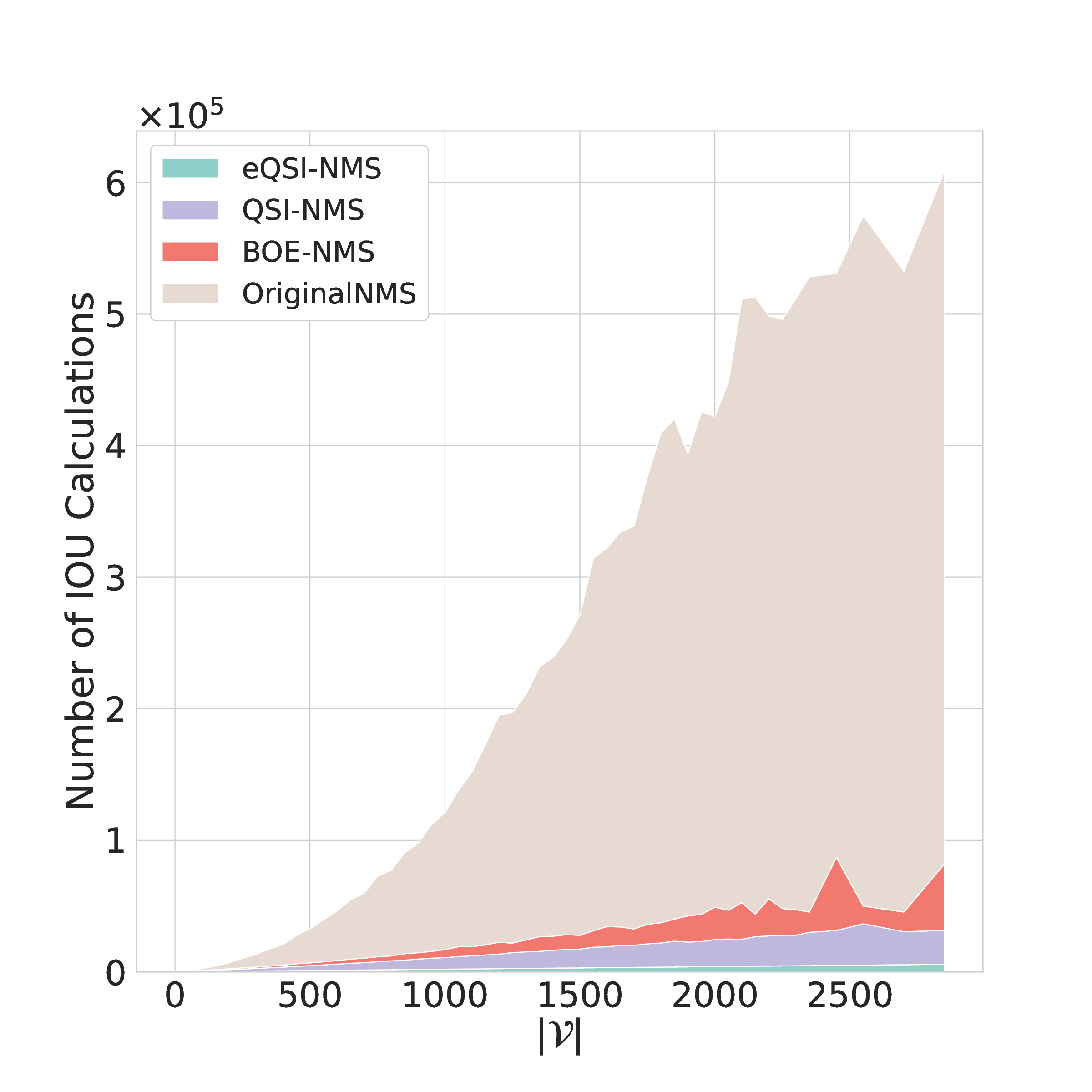}
        \end{minipage}
    }
    \subfigure[]{
        \label{sub2:b}
        \begin{minipage}{.48\linewidth}
        \includegraphics[width=\linewidth]{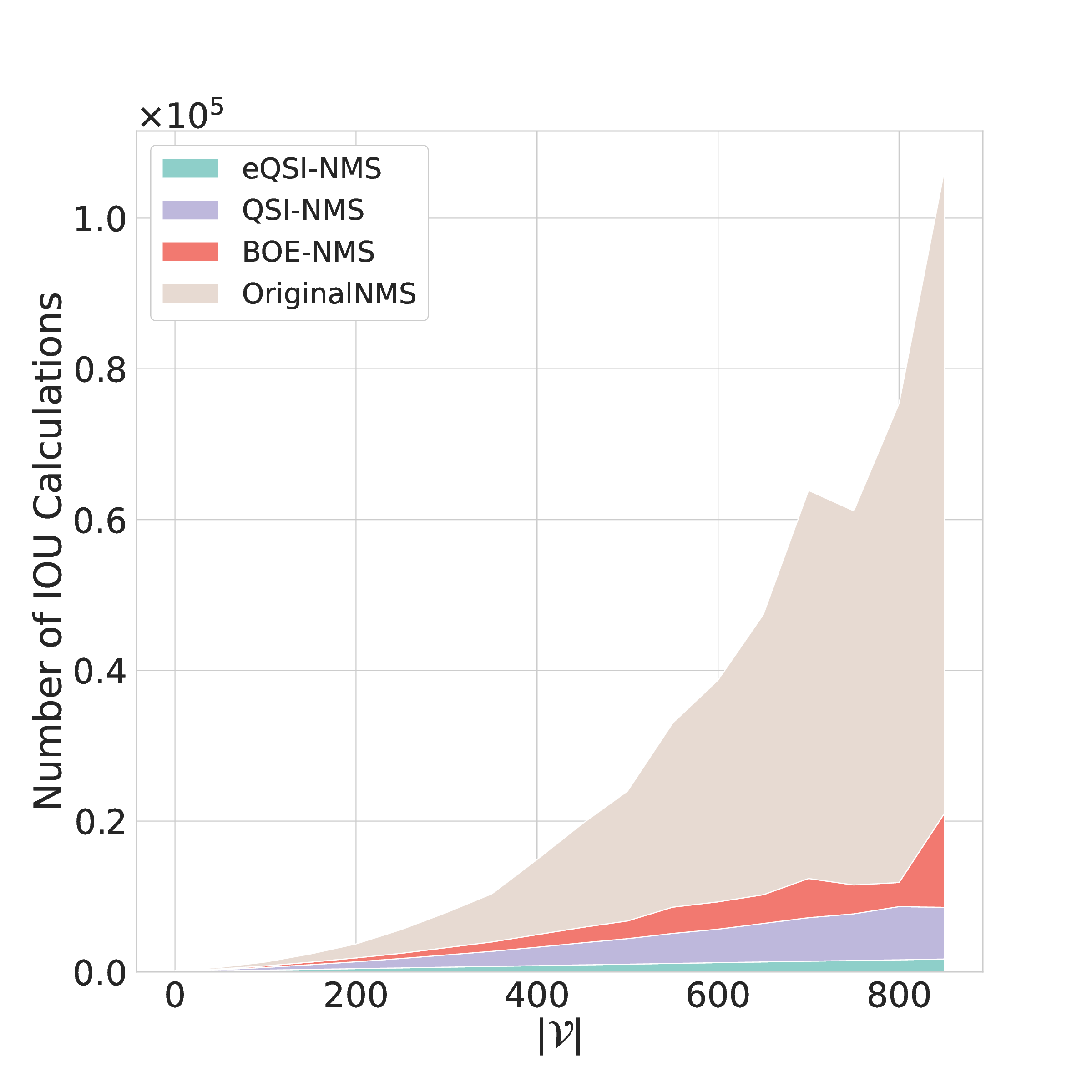}
        \end{minipage}
    }
    \caption{Stackplots of IOU calculations for different methods. \ref{sub2:a} shows the results of YOLOv8-N on MS COCO 2017, while \ref{sub2:b} shows the results of Faster R-CNN X101-FPN on MS COCO 2017.}
    \label{pic:appendix1}
\end{figure}

\paragraph{Statistics of IOU Calculations} During the NMS algorithm process, the computational cost of numerous IOU calculations is a performance bottleneck. We compare the number of IOU calculations between our methods and the original NMS. Figure~\ref{pic:appendix1} shows the relationship between the number of boxes and the number of IOU calculations for different methods. It can be observed that our methods significantly reduce the number of IOU calculations compared to original NMS, demonstrating the superiority.

\newpage
\paragraph{Results of Instance Segmentation Tasks} We also evaluate our methods on instance segmentation tasks using Mask R-CNN \cite{he2017mask} and YOLOv8, where they demonstrate significant superiority over other methods. Please refer to Table~\ref{tab:seg} for details.

\begin{table}[ht]
    \centering
    \caption{NMS Methods on Instance Segmentation Tasks}
    \label{tab:seg}
    \resizebox{\linewidth}{!}{
        \begin{tabular}{@{}cccccccccc@{}}
            \toprule
            \textbf{Model}           & \textbf{Size}      & \textbf{Target}          & \textbf{original NMS} & \textbf{Fast NMS} & \textbf{Cluster-NMS} & \textbf{BOE-NMS} & \textbf{QSI-NMS} & \textbf{eQSI-NMS} \\
            \midrule
            \multirow{3}{*}{Mask R-CNN R50-FPN}
&\multirow{2}{*}{-}&Average Latency ($\mu$s) &48.8&105.7&222.8&31.9&30.3&\textbf{21.6}\\
&& AP$^{\text{Box}}$ 0.5:0.95 (\%)&41.0&40.4&40.9&41.0&40.8&40.4\\
&& AP$^{\text{Mask}}$ 0.5:0.95 (\%)&37.2&36.9&37.1&37.2&36.9&36.7\\
            \multirow{3}{*}{Mask R-CNN R101-FPN}
&\multirow{2}{*}{-}&Average Latency ($\mu$s) &45.3&113.1&205.0&32.3&29.8&\textbf{21.5}\\
&& AP$^{\text{Box}}$ 0.5:0.95 (\%)&42.9&42.2&42.9&42.9&42.7&42.3\\
&& AP$^{\text{Mask}}$ 0.5:0.95 (\%)&38.6&38.4&38.6&38.6&38.4&38.2\\
            \multirow{3}{*}{Mask R-CNN X101-FPN}
&\multirow{2}{*}{-}&Average Latency ($\mu$s) &40.3&105.6&189.2&26.7&26.6&\textbf{19.3}\\
&& AP$^{\text{Box}}$ 0.5:0.95 (\%)&44.3&43.6&44.2&44.3&44.1&43.7\\
&& AP$^{\text{Mask}}$ 0.5:0.95 (\%)&39.5&39.3&39.5&39.5&39.3&39.1\\
            \midrule
            \multirow{6}{*}{YOLOv8}
&\multirow{3}{*}{N-SEG}&Average Latency ($\mu$s) &1265.3&366.4&859.4&219.5&153.4&\textbf{85.4}\\
&& AP$^{\text{Box}}$ 0.5:0.95 (\%)&36.7&36.5&36.7&36.7&36.6&36.4\\
&& AP$^{\text{Mask}}$ 0.5:0.95 (\%)&30.4&30.4&30.4&30.4&30.3&30.2\\
&\multirow{3}{*}{S-SEG}&Average Latency ($\mu$s) &740.0&269.2&736.2&158.6&115.8&\textbf{61.9}\\
&& AP$^{\text{Box}}$ 0.5:0.95 (\%)&	44.7&44.5&44.7&44.7&44.6&44.4\\
&& AP$^{\text{Mask}}$ 0.5:0.95 (\%)&36.7&36.6&36.7&36.7&36.5&36.5\\
            \bottomrule
        \end{tabular}
    }
\end{table}

\clearpage
\section{Pseudo-Codes}\label{pseudo-codes}
\subsection{Pseudo-Code for QSI-NMS}
\begin{algorithm}[h]
    \caption{QSI-NMS}\label{alg:qsi-nms}
    \SetAlgoLined
    % \SetKwData{Lowerbound}{LOWERBOUND}
    % \SetKwData{Upperbound}{UPPERBOUND}
    \SetKwInOut{Input}{Input}
    \SetKwInOut{Output}{Output}

    \SetKwFunction{QSINMS}{QSI-NMS}
    \SetKwFunction{Partition}{Partition}
    \SetKwProg{myproc}{Procedure}{}{}
    \SetKwProg{myfun}{Function}{}{}
     
    \Input{$\mathcal{B} = b_1,\cdots, b_n, \mathcal{C} = c_1, \cdots, c_n, \mathcal{S} = s_1, \cdots, s_n, N_t$\\
    $\mathcal{B}$ is the list of initial detection boxes, $\mathcal{C}$ contains the centroids of the boxes in $\mathcal{B}$, $\mathcal{S}$  contains corresponding detection scores, $N_t$ is the NMS threshold.}
   
    \Output{$\mathcal{D}:$ boxes to be retained.}
    % \myfun{\Partition{$l, r$}}{
    %     $\mathcal{S}' \leftarrow \mathcal{S}_{l\sim r}$\;
    %     $m \leftarrow \arg\max \mathcal{S}'$\;
    %     swap($b_m, b_r$); swap($s_m, s_r$); swap($s_m, s_r$)\;
    %     $p \leftarrow l$\;
    %     \For{$i\in [l, r]$} {
    %         \If{$c_i \preceq c_r$}{
    %             swap($b_p, b_i$); swap($s_p, s_i$); swap($c_p, c_i$)\;
    %             $p \leftarrow p + 1$\; 
    %         }
    %     }
    %     swap($b_p, b_r$); swap($s_p, s_r$); swap($c_p, c_r$)\;
    %     \Return{$p$}
    % }
    % \myproc{\QSINMS{$l, r,\delta$}}{
    %     $p\leftarrow$ \Partition{$l, r$}\; 
    %     \If{$\delta(b_p)$} {
    %         \For{$i\in [l, r]$} {
    %             \If{$\mathrm{IOU}(b_p, b_i) > N_t$}{
    %                 $\delta(b_p) \leftarrow \mathrm{False}$\;
    %             }
    %         }
    %         $\mathcal{D} \leftarrow \mathcal{D}\cup b_p$\;
    %     }
    %     \QSINMS{$l, p - 1, \delta$}\;
    %     \QSINMS{$p + 1, r, \delta$}\;
        
    % }
    \begin{minipage}{.5\textwidth}
        \myfun{\Partition{$l, r$}}{
            $\mathcal{S}' \leftarrow \mathcal{S}_{l\sim r}$\;
            $m \leftarrow \arg\max \mathcal{S}'$\;
            swap($b_m, b_r$); swap($s_m, s_r$); swap($c_m, c_r$)\;
            $p \leftarrow l$\;
            \For{$i\in [l,r-1]$} {
                \If{$c_i \preceq_{\mathcal{C}} c_r$}{
                    swap($b_p, b_i$)\; swap($s_p, s_i$)\; swap($c_p, c_i$)\;
                    $p \leftarrow p + 1$\; 
                }
            }
            swap($b_p, b_r$); swap($s_p, s_r$); swap($c_p, c_r$)\;
            \Return{$p$}
    }
    \end{minipage}%
    \begin{minipage}{.5\textwidth}
    \myproc{\QSINMS{$l, r,\delta$}}{
        \If{$l\geq r$}{
        \Return{}
        }
        $p\leftarrow$ \Partition{$l, r$}\; 
        \If{$\delta(b_p)$} {
            \For{$i\in [l, r]\setminus\{p\}$} {
                \If{$\mathrm{IOU}(b_p, b_i) > N_t$}{
                    $\delta(b_p) \leftarrow \mathrm{False}$\;
                }
            }
            $\mathcal{D} \leftarrow \mathcal{D}\cup b_p$\;
        }
        \QSINMS{$l, p - 1, \delta$}\;
        \QSINMS{$p + 1, r, \delta$}\;
    }
    \end{minipage}
    \Begin{
         $\mathcal{D} \leftarrow \{\}$;$\mathcal{\delta} \leftarrow \{\mathrm{True}\}^{n}$\;
        \QSINMS{$1, n, \delta$}\;
        \Return{$\mathcal{D}$}
    }
\end{algorithm}
\clearpage
\subsection{Pseudo-Code for eQSI-NMS}
\begin{algorithm}[ht]
    \caption{eQSI-NMS}\label{alg:eqsi-nms}
    \SetAlgoLined
    % \SetKwData{Lowerbound}{LOWERBOUND}
    % \SetKwData{Upperbound}{UPPERBOUND}
    \SetKwInOut{Input}{Input}
    \SetKwInOut{Output}{Output}

    \SetKwFunction{Solve}{Solve}
    \SetKwFunction{Partition}{Partition}
    \SetKwProg{myproc}{Procedure}{}{}
    \SetKwProg{myfun}{Function}{}{}
     
    \Input{$\mathcal{B} = b_1,\cdots, b_n, \mathcal{C} = c_1, \cdots, c_n, \mathcal{S} = s_1, \cdots, s_n, i_n, N_t$\\
    $\mathcal{B}$ is the list of initial detection boxes, $\mathcal{C}$ contains the centroids of the boxes in $\mathcal{B}$, $\mathcal{S}$  contains corresponding detection scores, $N_t$ is the NMS threshold.}
   
    \Output{$\mathcal{D}:$ boxes to be retained.}
  \myproc{\Solve{$I$}}{
        % Sort all elements of $\mathcal{C}$ in ascending order according to the partial order $\preceq_{\mathcal{C}}$ into a sequence $C$\;
        % $\mathcal{B}' \leftarrow \mathcal{B}$\;
         Stack $\mathcal{L}^{b} \leftarrow []$; $\mathcal{L}^{s} \leftarrow []$\;
          \For{$m=1, 2, \ldots, n$}{
          
         % $m \leftarrow \arg\min_{\preceq} \mathcal{C}$\;
         $b^{*} \leftarrow b_{I_m}$\;
         % $\mathcal{B}' \leftarrow \mathcal{B}' - b^*$\;
         \While{$\mathcal{L}^s$ is not empty $ \wedge$ $\mathcal{L}^b $ is not empty}{
            \If{$\mathrm{TOP}(\mathcal{L}^s)$ < $s_m$}{
            \If{$\mathrm{IOU}(b^*, \mathrm{TOP}(\mathcal{L}^b)) > N_t$}{
                $\delta(\mathrm{TOP}(\mathcal{L}^b)) \leftarrow \mathrm{False}$\;
            }
            $\mathrm{POP}(\mathcal{L}^b)$;$\mathrm{POP}(\mathcal{L}^s)$\;
            } \Else{
                \textbf{Break}\;
            } 
         }
           $\mathrm{PUSH}(\mathcal{L}^b,b^*)$; $\mathrm{PUSH}(\mathcal{L}^s, s_m)$\;
        }
    }
     
    \Begin{
         $\mathcal{D} \leftarrow \{\}$ ; $\mathcal{\delta} \leftarrow \{\mathrm{True}\}^{n}$\;
         $C \leftarrow$ the sorted $\mathcal{C}$ in ascending order according to $\preceq_{\mathcal{C}}$\;
         $I \leftarrow (i_1, i_2, \ldots, i_n)$ where $C = (c_{i_1}, c_{i_2}, \ldots, c_{i_n})$\;
        \Solve{I}\;
        $I \leftarrow$ reverse$(I)$\;
        \Solve{I}\;
        \For{$b \in \mathcal{B}$}{
            \If{$\delta(b)$}{
            $\mathcal{D}\leftarrow \mathcal{D} \cup b$\;
            }
        }
      
        \Return{$\mathcal{D}$}
        
    }
\end{algorithm}
\clearpage
\subsection{Pseudo-Code for BOE-NMS}
\begin{algorithm}[h]
    \caption{BOE-NMS}\label{alg:boe-nms}
    \SetAlgoLined
    % \SetKwData{Lowerbound}{LOWERBOUND}
    % \SetKwData{Upperbound}{UPPERBOUND}
    \SetKwInOut{Input}{Input}
    \SetKwInOut{Output}{Output}
    \Input{$\mathcal{B} = b_1,\cdots, b_n, \mathcal{M} = m_1, \cdots, m_n, \mathcal{S} = s_1, \cdots, s_n, \mathcal{I} = i_1, \cdots, i_n, N_t$\\
    $\mathcal{B}$ is the list of initial detection boxes, $\mathcal{M}$ contains the x-coordinates of the centroids of the boxes in $\mathcal{B}$, $\mathcal{S}$  contains corresponding detection scores, $\mathcal{I}$ contains the ranks of all boxes in $\mathcal{B}$ , which is sorted by x-coordinate of the centroids of the boxes in ascending order, $N_t$ is the NMS threshold.}
   
    \Output{$\mathcal{D}:$ boxes to be retained.}
    \Begin{
        $\mathcal{D} \leftarrow \{\}$\;
        \While{$\mathcal{B} \neq \emptyset$}{
            $m \leftarrow \arg\max \mathcal{S}$\;                    
            $b^{*} \leftarrow b_m$\;
            $\mathcal{D}\leftarrow\mathcal{D}\cup b^{*}$;$\mathcal{B}\leftarrow\mathcal{B}-b^{*}$;$\mathcal{S}\leftarrow\mathcal{S}-s_m$\;

            % TODO: color the difference of Greedy NMS and BOB NMS
            % \For{$b_i \in \mathcal{B}$}{
            %     \If{$\mathrm{iou}(\mathcal{M}, b_i) \geq N_t$}{
            %         $\color{red}{B\leftarrow B-b_i;S\leftarrow S-s_i}$\;
            %     }
            % }
            $x_l \leftarrow$ left x-coordinate$(b^{*})$;  $x_r \leftarrow$ right x-coordinate $(b^{*})$\;  
            $l \leftarrow $ lowerbound $(\mathcal{I}, x_l) $; \Comment{Find the rank of the first item i in $\mathcal{I}$, $\mathbf{s.t.} m_i \geq x_l$}\\
            $r \leftarrow $ upperbound $(\mathcal{I}, x_r)$; \Comment{Find the rank of the first item i in $\mathcal{I}$, $\mathbf{s.t.} m_i > x_r$}\\ 
            % TODO: explain for lowerbound and upperbound
            $\mathcal{I}' \leftarrow \mathcal{I}_{l\sim r-1}$ \;  
            \For{$i \in \mathcal{I}'$}{
                \If{$\mathrm{IOU}(b^{*}, b_i) > N_t$}{
                    $\mathcal{B}\leftarrow \mathcal{B}-b_i$; $\mathcal{S}\leftarrow \mathcal{S}-s_i$\;
                }
            }
        }
        \Return{$\mathcal{D}$}
    }
\end{algorithm}

\newpage
\section*{NeurIPS Paper Checklist}

\begin{enumerate}

\item {\bf Claims}
    \item[] Question: Do the main claims made in the abstract and introduction accurately reflect the paper's contributions and scope?
    \item[] Answer: \answerYes{} % Replace by \answerYes{}, \answerNo{}, or \answerNA{}.
    \item[] Justification: Please refer to Section~\ref{intro}, Paragraph 3, 4, and 5.

\item {\bf Limitations}
    \item[] Question: Does the paper discuss the limitations of the work performed by the authors?
    \item[] Answer: \answerYes{} % Replace by \answerYes{}, \answerNo{}, or \answerNA{}.
    \item[] Justification: Please refer to Appendix~\ref{discussion}, Paragraph 2.

\item {\bf Theory Assumptions and Proofs}
    \item[] Question: For each theoretical result, does the paper provide the full set of assumptions and a complete (and correct) proof?
    \item[] Answer: \answerYes{} % Replace by \answerYes{}, \answerNo{}, or \answerNA{}.
    \item[] Justification: Please refer to Appendix~\ref{pre} and \ref{proofs}.

    \item {\bf Experimental Result Reproducibility}
    \item[] Question: Does the paper fully disclose all the information needed to reproduce the main experimental results of the paper to the extent that it affects the main claims and/or conclusions of the paper (regardless of whether the code and data are provided or not)?
    \item[] Answer: \answerYes{} % Replace by \answerYes{}, \answerNo{}, or \answerNA{}.
    \item[] Justification: For algorithm implementation details, please refer to Section~\ref{methodology} and Appendix~\ref{pseudo-codes}. And for experimental details, please refer to Section~\ref{exp} and Appendix~\ref{exp-details}.

\item {\bf Open access to data and code}
    \item[] Question: Does the paper provide open access to the data and code, with sufficient instructions to faithfully reproduce the main experimental results, as described in supplemental material?
    \item[] Answer: \answerYes{} % Replace by \answerYes{}, \answerNo{}, or \answerNA{}.
    \item[] Justification: We have included the code, related documentation, and licenses in the supplementary material. We will open-source our code and data after acceptance. 

\item {\bf Experimental Setting/Details}
    \item[] Question: Does the paper specify all the training and test details (e.g., data splits, hyperparameters, how they were chosen, type of optimizer, etc.) necessary to understand the results?
    \item[] Answer: \answerYes{} % Replace by \answerYes{}, \answerNo{}, or \answerNA{}.
    \item[] Justification: Please refer to Section~\ref{exp} and Appendix~\ref{exp-details}. More details can be found in our source code.

\item {\bf Experiment Statistical Significance}
    \item[] Question: Does the paper report error bars suitably and correctly defined or other appropriate information about the statistical significance of the experiments?
    \item[] Answer: \answerYes{} % Replace by \answerYes{}, \answerNo{}, or \answerNA{}.
    \item[] Justification: Please refer to Table~\ref{tab:COCO}, Table~\ref{tab:OI}, Table~\ref{tab:tc}, and Figure~\ref{pic:appendix1}.

\item {\bf Experiments Compute Resources}
    \item[] Question: For each experiment, does the paper provide sufficient information on the computer resources (type of compute workers, memory, time of execution) needed to reproduce the experiments?
    \item[] Answer: \answerYes{} % Replace by \answerYes{}, \answerNo{}, or \answerNA{}.
    \item[] Justification: Please refer to Table~\ref{tab:expenv} for the CPU/GPU details.
    
\item {\bf Code Of Ethics}
    \item[] Question: Does the research conducted in the paper conform, in every respect, with the NeurIPS Code of Ethics \url{https://neurips.cc/public/EthicsGuidelines}?
    \item[] Answer: \answerYes{} % Replace by \answerYes{}, \answerNo{}, or \answerNA{}.
    \item[] Justification: The research conducted in this article fully complies with the NeurIPS Code of Ethics.

\item {\bf Broader Impacts}
    \item[] Question: Does the paper discuss both potential positive societal impacts and negative societal impacts of the work performed?
    \item[] Answer: \answerNA{} % Replace by \answerYes{}, \answerNo{}, or \answerNA{}.
    \item[] Justification: Our research on NMS is foundational work in the field of object detection, and does not have any societal impact.
    
\item {\bf Safeguards}
    \item[] Question: Does the paper describe safeguards that have been put in place for responsible release of data or models that have a high risk for misuse (e.g., pretrained language models, image generators, or scraped datasets)?
    \item[] Answer: \answerNA{} % Replace by \answerYes{}, \answerNo{}, or \answerNA{}.
    \item[] Justification: This paper does not present any such risks.

\item {\bf Licenses for existing assets}
    \item[] Question: Are the creators or original owners of assets (e.g., code, data, models), used in the paper, properly credited and are the license and terms of use explicitly mentioned and properly respected?
    \item[] Answer: \answerYes{} % Replace by \answerYes{}, \answerNo{}, or \answerNA{}.
    \item[] Justification: We cite the relevant models and datasets. Additionally, we provide the corresponding copyrights and licenses in the supplementary material.

\item {\bf New Assets}
    \item[] Question: Are new assets introduced in the paper well documented and is the documentation provided alongside the assets?
    \item[] Answer: \answerYes{} % Replace by \answerYes{}, \answerNo{}, or \answerNA{}.
    \item[] Justification: We provide the corresponding instructional documentation and the license in the supplementary material.

\item {\bf Crowdsourcing and Research with Human Subjects}
    \item[] Question: For crowdsourcing experiments and research with human subjects, does the paper include the full text of instructions given to participants and screenshots, if applicable, as well as details about compensation (if any)? 
    \item[] Answer: \answerNA{} % Replace by \answerYes{}, \answerNo{}, or \answerNA{}.
    \item[] Justification: This paper does not involve crowdsourcing nor research with human subjects.

\item {\bf Institutional Review Board (IRB) Approvals or Equivalent for Research with Human Subjects}
    \item[] Question: Does the paper describe potential risks incurred by study participants, whether such risks were disclosed to the subjects, and whether Institutional Review Board (IRB) approvals (or an equivalent approval/review based on the requirements of your country or institution) were obtained?
    \item[] Answer: \answerNA{} % Replace by \answerYes{}, \answerNo{}, or \answerNA{}.
    \item[] Justification: This paper does not involve crowdsourcing nor research with human subjects.

\end{enumerate}

\end{document}